\documentclass[11pt]{article}

\usepackage{acl}

\usepackage{times}
\usepackage{latexsym}

\usepackage[T1]{fontenc}

\usepackage[utf8]{inputenc}

\usepackage{microtype}

\usepackage{inconsolata}

\usepackage{graphicx}
\usepackage{multirow}
\usepackage{listings}
\usepackage[inkscapelatex=false]{svg}
\definecolor{commentgreen}{rgb}{0.0, 0.5, 0.0} \definecolor{stringpurple}{rgb}{0.58, 0.0, 0.82} \definecolor{keywordblue}{rgb}{0.0, 0.0, 1.0} 
\definecolor{darkblue}{rgb}{0.0, 0.0, 0.55}
\definecolor{backgray}{rgb}{1, 1, 1} 
\definecolor{red}{rgb}{1.0, 0.0, 0.0} 
\usepackage{pgf-pie} 
\usepackage{array,makecell} 
\newcommand{\data}{HistoryBank}
\newcommand{\task}{HistoryBankQA}
\title{\task{}: Multilingual Temporal Question Answering on Historical Events}

\author{Biswadip Mandal \\
  \texttt{biswadip.iitb@gmail.com} \\\And
  Anant Khandelwal \\
  \texttt{anantk@microsoft.com}   \\\And
   Manish Gupta \\
  \texttt{gmanish@microsoft.com}  \\}

\begin{document}
\maketitle
\begin{abstract}
Temporal reasoning about historical events is a critical skill for NLP tasks like event extraction, historical entity linking, temporal question answering, timeline summarization, temporal event clustering and temporal natural language inference. Yet efforts on benchmarking temporal reasoning capabilities of large language models (LLMs) are rather limited. Existing temporal reasoning datasets are limited in scale, lack multilingual coverage and focus more on contemporary events. 
To address these limitations, we present \data{}, a multilingual database of 10M+ historical events extracted from Wikipedia timeline pages and article infoboxes. Our database provides unprecedented coverage in both historical depth and linguistic breadth with 10 languages. Additionally, we construct a comprehensive question answering benchmark for temporal reasoning across all languages. This benchmark covers a diverse set of 6 temporal QA reasoning tasks, and we evaluate a suite of popular language models (LLaMA-3-8B, Mistral-7B, Gemma-2-9b, Qwen3-8B, GPT4o) to assess their performance on these tasks. As expected GPT4o performs best across all answer types and languages; Gemma-2 outperforms the other small language models. Our work aims to provide a comprehensive resource for advancing multilingual and temporally-aware natural language understanding of historical events. To facilitate further research, we will make our code and datasets publicly available upon acceptance of this paper.
\footnote{Code available at: \url{https://anonymous.4open.science/r/history-bank-4377}\label{codeDataFN}} 
\end{abstract}

\section{Introduction}
Temporal reasoning is the ability to understand, represent, and manipulate time-related information. Temporal reasoning about historical events is fundamental to a wide range of natural language processing (NLP) and knowledge-based applications, including event extraction, temporal question answering (QA), timeline generation, historical entity linking, temporal event clustering, timeline summarization, temporal natural language inference, planning, and narrative comprehension.
Accurate event extraction requires identifying and timestamping events in context, which demands a deep understanding of historical timelines and their nuances. 

\begin{table*}[!ht]
    \centering
    \scriptsize
    \begin{tabular}{|l|p{0.1\textwidth}|p{0.6\textwidth}|p{0.18\textwidth}|}
    \hline
\multicolumn{2}{|c|}{Question Type}&Question&Answer\\
\hline
\hline
\multicolumn{2}{|c|}{FactQA}&I'm looking for the date of the release of the Goodnight Punpun Omnibus 1. When was it released?&\{'year': '2016', 'month': '03', 'day': '04'\}\\
\hline
\multicolumn{2}{|c|}{DurationQA}&What was the time duration between the release of the single ``Happy Happy'' and the release of the single ``Fake \& True''?&start\_date=\{'year': '2019', 'month': '06', 'day': '12'\}, end\_date=\{'year': '2019', 'month': '10', 'day': '18'\}\\
\hline
\multirow{27}{*}{\rotatebox{90}{RelationQA}}&Compare Duration&Which event span is of longer duration: ``Agnes van Ardenne was born'' to ``Agnes van Ardenne started her earlier term as Member of the House of Representatives'', or ``Agnes van Ardenne ended her earlier term as Member of the House of Representatives'' to ``Agnes van Ardenne started her term as Member of the House of Representatives''?&``Agnes van Ardenne was born'' to ``Agnes van Ardenne started her earlier term as Member of the House of Representatives''.\\
\cline{2-4}
&Duration Diff&What is the difference in duration between ``A new session of the Alabama Legislature began'' to ``The next election for the Alabama Senate is scheduled'', and ``Anthony Daniels was elected as House Minority Leader of Alabama'' to ``The last election for the Alabama Senate was held''. (in days) in the context of ``Alabama Legislature''?&762\\
\cline{2-4}
&Gap Between&What is the gap in days between ``José Luis Soro became the leader of Chunta Aragonesista'' to ``Maru Díaz became the leader of Podemos–Green Alliance in Aragon.'' and ``Tomás Guitarte became the leader of Teruel Existe'' to ``Alberto Izquierdo became the leader of the Aragonese Party.'', in the context of ``2023 Aragonese regional election''?&1522\\
\cline{2-4}
&Inclusion&Which event's time span includes the other: ``June 2009 Extra Session of the 99th Wisconsin Legislature ended'' to ``December 2009 Special Session of the 99th Wisconsin Legislature began'', or ``Election for the 99th Wisconsin Legislature was held'' to ``The term of the 99th Wisconsin Legislature ended.'', in the context of ``99th Wisconsin Legislature''?&Election for the 99th Wisconsin Legislature was held to The term of the 99th Wisconsin Legislature ended.\\
\cline{2-4}
&Order Span End&Which event span ended last: ``The album `100 Reasons to Live' by Gareth Emery was released'' to ``The single `Far From Home' by Gareth Emery feat. Gavrielle was released'', or ``The single `Hands' by Gareth Emery \& Alastor feat. London Thor was released'' to ``The single `Save Me' by Gareth Emery was released.'', in the context of ``100 Reasons to Live''?&``The single `Hands' by Gareth Emery \& Alastor feat. London Thor was released'' to ``The single `Save Me' by Gareth Emery was released.''\\
\cline{2-4}
&Order Span Start&Which event span started earlier: ``The album `100 Reasons to Live' by Gareth Emery was released'' to ``The single `Far From Home' by Gareth Emery feat. Gavrielle was released'', or ``The single `Hands' by Gareth Emery \& Alastor feat. London Thor was released'' to ``The single `Save Me' by Gareth Emery was released.'', in the context of ``100 Reasons to Live''?&``The single `Hands' by Gareth Emery \& Alastor feat. London Thor was released'' to ``The single `Save Me' by Gareth Emery was released.''\\
\cline{2-4}
&Overlap&How many days do ``The single `Hands' by Gareth Emery \& Alastor feat. London Thor was released'' to ``The single `Far From Home' by Gareth Emery feat. Gavrielle was released.'', and ``The album `100 Reasons to Live' by Gareth Emery was released'' to ``The single `Save Me' by Gareth Emery was released.'' overlap, in the context of ``100 Reasons to Live''?&29\\
\hline
\multicolumn{2}{|c|}{CountQA}&The following is a list of historical events. Each line includes a description and the title of the article it comes from. (1) Event about `Sean Kingston (album)': The single ``Beautiful Girls'' by Sean Kingston was released. (2) Event about `Week End (X Japan song)': The song ``Week End'' by X Japan was released. (3) Event about `Lucy McEvoy': Lucy McEvoy was nominated for the AFL Women's Rising Star award. (4) Event about `BL 5.4-inch howitzer': The Ordnance BL 5.4-inch howitzer was used in the Second Boer War. (5) Event about `Holten Castenschiold': Holten Castenschiold began his term as the 6th President of the Danish Olympic Committee. (6) Event about `William L. Baird': William Lewis Baird began his term as the 19th Mayor of Lynn, Massachusetts. (7) Event about `Jung Yeon-kyung': Jung Yeon-kyung was born. Provide the count of the number of events that occurred during the 19th century. &2\\
\hline
\multicolumn{2}{|c|}{SequenceQA}&Sort the events by the time they took place. Each event is accompanied by a description and the title of the article it comes from. Return the correct chronological order by listing the event numbers, like (2) (1) (3). (1) Event about `Paaliaq': Paaliaq was discovered. (2) Event about `Bob Chakales': Bob Chakales was born. (3) Event about `Dark Valley': The film ``Dark Valley'' was released.&(2) (3) (1)\\
\hline
\multicolumn{2}{|c|}{RecurrenceQA}&The following event includes the article title and event description. Event about `2008 Iranian legislative election': The second round of the 2008 Iranian legislative election was held. Identify the year when the previous edition of the event took place. Provide the year as your answer.&2004\\
\hline
    \end{tabular}
    \caption{Examples of English Questions from \task{}. More examples are in Tables~\ref{tab:questions_english_part1},~\ref{tab:questions_english_part2} and~\ref{tab:questions_english_part3} in Appendix~\ref{app:examples}.}
    \label{tab:examples}
\end{table*}

Temporal QA involves answering questions like ``Who was the president during WWII?'', which necessitates reasoning over historical periods and contextual knowledge. Timeline generation builds coherent narratives by organizing events chronologically and interpreting their temporal relationships. Historical entity linking ensures that mentions like ``King George'' are correctly resolved to the appropriate historical figure, relying heavily on temporal disambiguation. Further, temporal event clustering groups events based on their temporal proximity or shared historical context, aiding in pattern discovery and narrative coherence. Timeline summarization, particularly in complex social media narratives \cite{song2024temporal}, condenses long historical narratives into structured, time-ordered summaries, making complex temporal data more digestible. Finally, temporal natural language inference enables systems to reason about temporal entailment and contradiction, such as understanding whether one event logically follows or contradicts another based on time. Measuring and improving temporal natural language inference is not only a theoretical exercise but has broad practical implications for information synthesis and reasoning.

Accurately benchmarking large language models (LLMs) for their temporal reasoning capabilities presents two key challenges. First, there is a need for a large, multilingual corpus of historical events that captures diverse temporal contexts across cultures and time periods. Second, designing a comprehensive benchmark that effectively evaluates an LLM’s ability to perform a wide range of temporal reasoning tasks (such as temporal entailment, ordering, duration inference, and temporal question answering) requires leveraging this corpus to create rich, diverse, and challenging test scenarios.

Towards the first challenge, efforts have been made to extract events from the open web, social media, and textual corpora. For instance, prior work has explored extracting public events from unstructured web sources \cite{wang2019constructing}, mining local social activities from Facebook pages \cite{chang2020eventgo}, and developing document-level event extraction systems driven by heuristically defined prompts \cite{liu2024document}. More recently, researchers have begun to tackle multilingual event extraction from low-resource historical corpora such as colonial newspaper advertisements \cite{borenstein2023multilingual}, which presents unique challenges due to linguistic variation, annotation scarcity, and domain shift. These advances have enabled the creation of dynamic, localized, and sometimes multilingual event databases, yet they remain largely focused on ephemeral, contemporary, or socially-driven content.

In contrast, historical events, spanning centuries and encompassing political, scientific, cultural, and social milestones, are underrepresented in current event resources. This is a critical gap: historical events not only encode encyclopedic knowledge but also offer a temporally rich substrate for modeling factual consistency, event progression, and cause-effect relations over long time horizons. Such structured historical knowledge is essential for downstream tasks like timeline summarization, temporal QA, and the evaluation of LLMs' ability to reason about time in grounded, factual contexts. However, the construction of high-quality, temporally anchored historical event datasets remains a challenging and underexplored area. Unlike contemporary event mining from dynamic web content, assembling a comprehensive and multilingual historical event database requires synthesizing semi-structured sources, resolving temporal ambiguity, and aligning event granularity across languages.

Hence, to address the first challenge, we propose a large-scale multilingual event database, \data{}. We construct an extensive repository of dated events by extracting structured information from event-centric Wikipedia pages.
Specifically, we integrate data from \emph{On This Day} pages\footnote{For example, for English, this page links to a page per day of the year: \url{https://en.wikipedia.org/wiki/Category:Days_of_the_year}.}, which enumerate historical events for each calendar date, and from event-centric infoboxes, which encode temporal metadata associated with entities. 
The resulting database spans ten languages: English (en), Bengali (bn), German (de), French (fr), Indonesian (id), Hindi (hi), Italian (it), Portuguese (pt), Russian (ru), and Spanish (es) and captures events across diverse domains including politics, culture, science, and sports. For English alone, the corpus comprises $\sim$8.2M event records, representing the largest publicly available collection of multilingual temporal event data. 

Towards the second challenge, recent benchmarks, such as TRAM \cite{wang-zhao-2024-tram}, TG-LLM \cite{xiong-etal-2024-large}, TempReason \cite{tan2023towards}, and ChronoSense \cite{islakoglu2025chronosense}, have explored LLMs' temporal reasoning abilities, though most focus on synthetic setups, abstract events, or everyday temporal phenomena rather than historically grounded contexts.

In contrast, our work focuses on temporal reasoning over curated, encyclopedic historical events, factual data grounded in time and reality. This setting allows us to simultaneously evaluate both memorization (i.e., factual recall) and temporal reasoning (e.g., ordering, comparison, disambiguation) in a unified framework. We argue that this dual challenge better approximates real world usage, where users expect models to reason over temporally structured factual knowledge, not just toy examples or synthetic timelines. Moreover, our multilingual design, spanning ten typologically diverse languages, enables the investigation of temporal reasoning across linguistic and cultural boundaries, a setting still underexplored in the current literature. 

We built six temporal QA tasks, together called as \task{},  leveraging the event corpus. They encompass a range of temporal phenomena, including (i) explicit date retrieval (FactQA), (ii) event sequencing and ordering (SequenceQA), (iii) temporal interval computation (DurationQA), (iv) relative time expression resolution (RelationQA), (v) counting number of events that occurred in a century or between two events/dates (CountQA), and (vi) identifying previous or next occurrences of a recurring event (RecurrenceQA). Each dataset is constructed to systematically probe model performance on well-defined temporal reasoning tasks across multiple languages and event types. Table~\ref{tab:examples} shows an English question example from our dataset for each question type. More examples are in Tables~\ref{tab:questions_english_part1},~\ref{tab:questions_english_part2} and~\ref{tab:questions_english_part3} in Appendix~\ref{app:examples}.

By combining large-scale multilingual event representations with diverse, carefully curated QA tasks, our resources facilitate rigorous evaluation of temporal reasoning in LLMs and fine-tuned models. We will release the code, event database and QA benchmarks upon acceptance of this paper, 
\footref{codeDataFN} 
to advance research into temporal understanding, support reproducible evaluation protocols, and enable the development of models with enhanced temporal inference capabilities.

Overall, we make the following main contributions in this paper.
\begin{itemize} \item Dataset Contribution: We introduce \data{}, the largest publicly available multilingual historical event database to date with 10M+ events across 10 languages, constructed using Wikipedia's On This Day pages and event-centric infoboxes. 
\item Benchmark Contribution: We design a comprehensive suite of six temporal question answering (QA) tasks that systematically evaluate different dimensions of temporal reasoning. These tasks include explicit date retrieval, event ordering, interval computation, relative time resolution, temporal comparison, and scope disambiguation. Each task is multilingual and grounded in real-world historical events, enabling robust cross-lingual and cross-domain evaluation of temporal understanding in LLMs.
\item Baseline Evaluation: We provide initial baseline results using both small language models and GPT4o under zero-shot settings. These results highlight the strengths and limitations of current models in handling temporally grounded factual reasoning, and establish a foundation for future research in multilingual temporal inference.
\end{itemize}

\section{Related Work}
Structured representations of historical events are foundational for temporal reasoning and knowledge-based tasks. Several projects have aimed to extract event data from semi-structured or unstructured sources. For example, GDELT \cite{leetaru2013gdelt} compiles large-scale political events from news sources for real-time monitoring, while EventKG \cite{gottschalk2018eventkg} builds a multilingual knowledge graph of events by aligning information across structured knowledge bases like Wikidata \cite{vrandevcic2014wikidata} , DBpedia \cite{auer2007dbpedia}, and YAGO \cite{hoffart2013yago2}. However, these efforts either focus on contemporary geopolitical events or rely heavily on existing ontologies. Our approach differs in that it aggregates historically anchored events directly from community-curated Wikipedia structures (timeline pages and infoboxes), yielding fine-grained, dated events across centuries and domains. Moreover, our focus on multilinguality and diversity of domains (science, culture, politics) distinguishes our resource from event datasets that are either monolingual or domain-restricted.

Temporal reasoning has recently gained attention in the area of evaluation of LLMs. TRAM \cite{wang-zhao-2024-tram} introduces a suite of datasets covering aspects such as event ordering, arithmetic, duration, and frequency, revealing significant performance gaps between current models and human baselines. But TRAM relies largely on abstract or synthetic scenarios. Other work has explored graph-based reasoning frameworks (e.g., TG-LLM \cite{xiong-etal-2024-large}) and multi-level temporal QA datasets such as TempReason \cite{tan2023towards}, aiming to uncover and enhance the temporal logic abilities of LLMs. TempReason \cite{tan2023towards} proposes a probing dataset organized by reasoning complexity but does not incorporate multilingual or real-world event contexts. TG-LLM \cite{xiong-etal-2024-large} enhances reasoning by introducing temporal graph representations.  ChronoSense \cite{islakoglu2025chronosense} further highlights the difficulty LLMs face in applying Allen's~\cite{allen1983maintaining} interval relations to both abstract and Wikidata-based events.

While these benchmarks offer valuable insights, they are primarily grounded in synthetic setups, abstract events, or everyday temporal phenomena. In contrast, our benchmark is grounded in factual, historical events with real-world time stamps, allowing joint assessment of memorization and reasoning. This setup better reflects the kinds of temporally grounded queries LLMs encounter in practice. 

\section{\data{}: A Large-scale Multilingual Database of Historical Events}
\label{sec:events}
We construct a multilingual event database from the Wikipedia dump by extracting structured events from infoboxes. Infoboxes are semi-structured templates found on Wikipedia pages that summarize salient facts about an entity using key-value pairs. Many of these values are temporally grounded, such as birth dates, foundation years, or historical milestones, which can be interpreted as implicit events.

\begin{table}[ht]
\centering
\scriptsize
\tabcolsep3pt
\begin{tabular}{|l|c|c|c|p{1.25cm}|p{1.25cm}|}
\hline
Language & Articles & Infoboxes & Events & Events per Infobox & Events per Article  \\
\hline \hline
English     & 4,260,757 & 4,729,733 & 8,201,400 & 1.73 & 1.92 \\
Bengali     & 47,350    & 52,166    & 78,181    & 1.50 & 1.65 \\
German      & 998,020   & 1,085,679 & 1,519,208 & 1.40 & 1.52 \\
French      & 113,489   & 199,910   & 112,207   & 0.56 & 0.99 \\
Indonesian  & 314,856   & 346,912   & 66,096    & 1.90 & 2.10 \\
Hindi       & 50,946    & 59,885    & 162,717   & 2.71 & 3.19 \\
Italian     & 6,548     & 6,766     & 19,473    & 2.87 & 2.97 \\
Portuguese  & 10,774    & 12,846    & 10,201    & 0.79 & 0.95 \\
Russian     & 5,756     & 6,915     & 7,979     & 1.15 & 1.39 \\
Spanish     & 8,688     & 19,847    & 23,536    & 1.18 & 2.71 \\
\hline
\end{tabular}
\caption{Language-wise \data{} Statistics}
\label{tab:events_stats}
\end{table}

Our key insight is that these temporally annotated infobox entries can be transformed into a rich source of historical and entity-centric events. For instance, the infobox of \textit{Richard Nixon} may contain an entry like \texttt{born = January 9, 1913}, which we convert into the event: ``Richard Nixon was born on January 9, 1913.''

To extract such events, we apply a multilingual language model (GPT4o) to each infobox, prompting it to:
\begin{itemize}
    \item Identify key-value pairs that contain date expressions.
    \item Parse the date into structured components (year, month, day).
    \item Generate a brief, natural language description of the event.
    \item Record the original key-value source.
    \item Annotate all countries potentially relevant to the event.
    \item Assign a country-specific polarity label (positive, negative, neutral).
\end{itemize}

This extraction pipeline is applied across ten languages: English, Bengali, German, French, Indonesian, Hindi, Italian, Portuguese, Russian, and Spanish. We prioritize infoboxes over free-text content due to their structural regularity, which allows for more reliable parsing and interpretation by LLMs. The complete prompt is listed in Appendix \ref{appendix:event_extraction_prompt}. Some of the automatically generated event descriptions are overly generic and lack sufficient context to be meaningful in isolation. For instance, a description like ``The individual played for the Oakland Raiders'' does not specify who the individual is. To address this limitation and improve clarity, we prepend the article title to each event description.

Table~\ref{tab:events_stats} summarizes the resulting dataset’s scale. English alone yields over 8 million events extracted from 4.7 million infoboxes, with an average of 1.73 events per infobox. Other languages follow with varying density, contributing to a diverse and scalable multilingual event corpus. The significant variation in events per infobox and events per article underscores the differing conventions across languages in how infoboxes are employed.

\begin{figure}[htbp]
    \centering
    \includegraphics[width=0.8\linewidth]{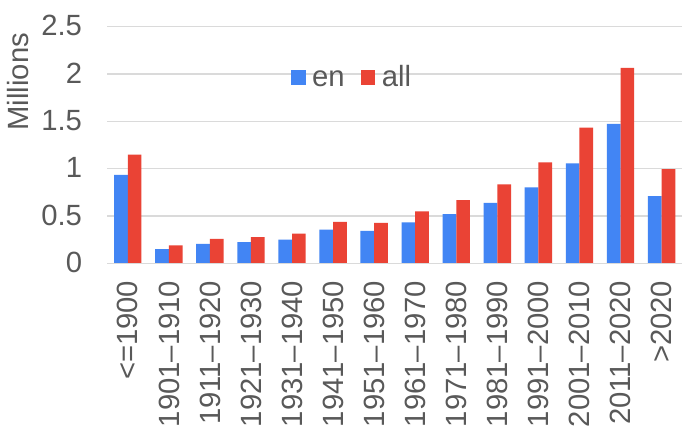}
    \caption{Time-wise Distribution of Events (English and all) in \data{}.}
    \label{fig:myplot}
\end{figure}

Figure \ref{fig:myplot} presents a distribution of when the events occurred for English events as well as all events from \data{}. The data shows a consistent rise in the number of recorded events over time, with a sharp acceleration starting from the 1960s. English-language events closely mirror the overall trend. The dataset contains a sizeable number of events before 1900; there are 11687 events even before 0 (BC). Appendix~\ref{app:detailedDatasetAnalysis} presents further detailed analysis of the dataset. Fig.~\ref{fig:eventDistributionAllLangs} shows time-wise distribution of events across all languages in \data{}. This also follows the same pattern as English. Further, we also how the distribution of English events across various entity (i.e., infobox) types in Fig.~\ref{fig:infoboxTypeDistribution-en}. We show most frequent 50 infobox types for events from other languages in Table~\ref{tab:infoboxTypeDistribution-all}. Table~\ref{tab:countryList} presents a list of most frequent 50 countries related to the events across languages. Fig.~\ref{fig:polarityDist} shows polarity distribution of events across all languages in \data{} dataset.


We also included an additional dataset sourced from Wikipedia’s On This Day pages. To collect this data, we automatically downloaded and organized historical event information (about births, deaths and generic events) from Wikipedia’s day-specific pages. The extracted information was standardized by associating each entry with a list of relevant countries and a polarity label. Table~\ref{tab:onthisday} in Appendix~\ref{app:detailedDatasetAnalysis} shows data statistics per language.

\begin{table*}[ht]
\centering
\scriptsize
\begin{tabular}{|l|l|c|c|c|c|c|c|c|c|c|c|}
\hline
\multicolumn{2}{|c|}{\textit{\textbf{Task}}} & English & Bengali & German & French & Indonesian & Hindi & Italian & Portuguese & Russian & Spanish \\
\hline \hline
\multicolumn{2}{|c|}{FactQA}& 4413790 & 38479   & 765258 & 70774  & 468121     & 86814 & 13341   & 4842       & 3888    & 19877   \\
\multicolumn{2}{|c|}{DurationQA}& 1731411 & 5549    & 30859  & 2250   & 47948      & 7328  & 312     & 123        & 482     & 308     \\ 
\hline
\multirow{7}{*}{\rotatebox{90}{RelationQA}}&Compare Duration                   & 1243222 & 5789    & 5460   & 8583   & 28677      & 7178  & 46      & 46         & 20      & 64      \\
&Duration Diff                      & 1449966 & 5782    & 5457   & 8532   & 28506      & 7098  & 46      & 46         & 20      & 64      \\
&Gap Between                        & 873727  & 684     & 760    & 1611   & 4344       & 1210  & 6       & 7          & 5       & 16      \\
&Inclusion                          & 839775  & 1219    & 583    & 1314   & 7193       & 1449  & 1       & 4          & 9       & 9       \\
&Order Span End                     & 2120530 & 2936    & 1700   & 3766   & 13902      & 3623  & 9       & 11         & 12      & 32      \\
&Order Span Start                   & 2120530 & 2936    & 1700   & 3766   & 13902      & 3623  & 9       & 11         & 12      & 32      \\
&Overlap                            & 1742192 & 5124    & 4707   & 6924   & 24718      & 6058  & 40      & 42         & 15      & 48      \\ \hline
\multicolumn{2}{|c|}{CountQA}& 1500000 & 14000   & 270000 & 20000  & 12000      & 29000 & 3500    & 1850       & 1450    & 4200    \\
\multicolumn{2}{|c|}{SequenceQA}& 2100000 & 19545   & 379802 & 28052  & 16524      & 40679 & 4868    & 2550       & 1995    & 5884    \\
\multicolumn{2}{|c|}{RecurrenceQA}& 42620   & 172     & 0      & 0      & 2243       & 258   & 0       & 0          & 0       & 0  \\
\hline
\end{tabular}
\caption{Number of examples per temporal reasoning task and language in our proposed \task{} benchmark.}
\label{tab:qa-dataset-stats}
\end{table*}

\section{\task{}: Temporal QA Benchmark}
One of the best ways to evaluate temporal reasoning ability of LLMs is to check their performance on temporal QA tasks. Our multilingual benchmark suite, \task{}, comprises of six distinct temporal question answering (QA) tasks, each targeting a specific temporal reasoning capability. All of these task datasets are derived from the multilingual event database introduced in Section~\ref{sec:events}, and follow a consistent schema across ten languages. In this section, we describe the construction methodology for the six core tasks: \textit{FactQA}, \textit{DurationQA}, and \textit{RelationQA}, \textit{SequenceQA}, \textit{CountQA}, \textit{RecurrenceQA}. Table~\ref{tab:qa-dataset-stats} shows the number of examples per temporal reasoning task and language in our proposed \task{} benchmark.  

The factual question answering task (FactQA) assesses a model’s capability to identify or deduce the specific date on which a given event took place, based on a natural language question.
Sequencing tasks (SequenceQA) assess the ability to order events chronologically, while duration tasks (DurationQA) focus on determining how long an event lasted. Relation labeling tasks (RelationQA) evaluate whether a model can identify temporal relationships between events, such as whether one event occurred before, after, or during another. Comparative tasks test whether a model can reason about relative durations, for example, by determining if one event lasted longer than another. 

Counting tasks (CountQA) require identifying the number of events that occurred between two time points or within a specified interval.
Recurrence question answering task (RecurrenceQA) focuses on recurring events such as elections, sports leagues, or international tournaments. Here, the goal is to assess whether a model understands periodicity by answering questions about the timing of previous iterations of a recurring event. These tasks aim to capture coarse-grained temporal understanding and fill gaps in existing evaluation paradigms.

\subsection{FactQA: Factual Date Identification}

FactQA evaluates a model's ability to retrieve or infer the date on which a specific event occurred, given a natural language question. For each extracted event in our database, we use its associated title, event description, and temporal key to construct a self-contained context. Using this context, we prompt an instruction-tuned generative language model (RecurrentGemma-2B-IT \cite{botev2024recurrentgemma} for English and Gemma-3B-IT \cite{team2025gemma} for other languages) to generate a natural question that asks for the timing of the event.

This generation pipeline is applied uniformly across all ten languages, ensuring linguistic diversity and contextual fidelity. The prompt is listed in Appendix~\ref{appendix:fact_qa_prompt}.

\subsection{DurationQA: Temporal Span Reasoning}

DurationQA requires the model to infer the temporal duration between two related events associated with the same entity. We first identify events in the same article. We then sort the events chronologically and form possible start-end pairs of events. To further filter out the valid start-end event pairs, we prompt a lightweight model (Flan-T5-base \cite{chung2024scaling}) to classify if the event pair is valid.  Filtered pairs are then passed to the question generation step, where we prompt a generative model (RecurrentGemma-2B-IT / Gemma-3B-IT) to write specific questions asking about the time elapsed between the two events in the context of the entity.
Prompts are included in Appendix~\ref{app:durationQA}.

\subsection{RelationQA: Temporal Span Comparison}

RelationQA assesses a model's ability to perform comparative reasoning over pairs of event spans, such as identifying which event lasted longer, whether they overlapped, or which began earlier. For each article with at least two distinct temporal spans (or durations), we generate all unique unordered pairs of event spans. For each pair, we compute derived temporal properties: duration, overlap, inclusion, gap, and relative ordering. Spans with identical start and end points are excluded.

Each event span pair is annotated with up to seven question types, designed to capture different aspects of comparative temporal reasoning:

\begin{enumerate}
\item Compare Duration: These questions assess which of the two spans lasted longer.
\item Duration Difference: These questions ask for the absolute difference in duration between the two spans.
\item Overlap: These questions measure the extent to which the two spans overlap, typically in terms of the number of overlapping days.
\item Gap Between: These questions determine the number of days separating the two spans, if any.
\item Inclusion: These questions evaluate whether one span is entirely contained within the other.
\item Order Start: These questions identify which span began earlier.
\item Order End: These questions identify which span concluded later.
\end{enumerate}

Each question is generated using natural language structure templates anchored on event descriptions and entity context. Answer keys are automatically derived from the known dates. By incorporating multiple reasoning categories under a single framework, RelationQA offers a fine-grained testbed for comparative temporal reasoning in language models.

\subsection{SequenceQA: Chronological Ordering Of Events}
SequenceQA is a QA task focused on the correct sequencing of historical events. This task captures three primary temporal reasoning sub-task types, each framed as a distinct QA format.

\begin{enumerate}
    \item Sequence Arrangement (Freeform Order Prediction): Models are given a set of unordered events and asked to arrange them in chronological order. Task is described using this example sentence: ``Arrange the following events in the order they occurred.''
   \item Multiple Choice Question (MCQ): Models are presented with multiple candidate sequences and must choose the one that is temporally accurate. Task is described using this example sentence: ``Which of the following sequences places the events in the right chronological order?''
   \item Order Verification (True/False): A single sequence of events is provided, and models must determine whether the sequence is chronologically correct. Task is described using this example sentence: ``Is this the correct historical order of the events?''
\end{enumerate}

To avoid any structural biases and encourage generalization, each task description sentence was diversified using 10 different natural language syntactic variant question templates. 





\subsection{CountQA: Counting Events}
The CountQA dataset consists of question-answer pairs that require models to reason about the number of historical events within specific temporal bounds. For each question, we sample 4 to 7 events, and . 
design three primary types of counting questions as follows.
\begin{enumerate}
    \item Count Between Events: Given a sequence of events (not necessarily in chronological order), the question asks how many events fall chronologically between two specific events in the list.
    Example: ``How many events occurred between event (2) and event (5)?''
    \item Count Between Years: A time range is specified using two randomly sampled years ending in 00 (e.g., 1200-1400), and the model must count how many events from the list fall within that range.
    Example: ``How many of the events occurred between the years 1500 and 1700?''
    \item Count by Century: The model is asked to count how many events occurred in a given century (e.g., 19th century). Centuries are sampled according to their observed frequency distribution in the data to ensure realistic coverage.
    Example: ``How many events below occurred in the 20th century?''
\end{enumerate}
   
Each question is generated using natural language paraphrase templates to introduce diversity in phrasing while maintaining clarity.

\subsection{RecurrenceQA : Identifying Previous Occurrences Of Recurrent Events}
To systematically evaluate temporal reasoning over recurring events, we construct a dataset consisting of questions that require identifying the year when a prior occurrence of a given event occurred. Candidate events are first identified by selecting entries whose metadata indicates a temporal relationship to an earlier occurrence. For example, if an event record (e.g., \url{https://en.wikipedia.org/wiki/2011_Cricket_World_Cup}) explicitly refers to a previous year or season (e.g., \url{https://en.wikipedia.org/wiki/2007_Cricket_World_Cup}), we interpret this reference as evidence of recurrence. To validate these relationships, we retrieve the corresponding current version of the event within the same article group and verify that the annotated date fields are well-formed and temporally consistent; the year of the current event must exceed the year of the referenced prior event. This validation step ensures that the resulting questions are anchored in correct chronological sequences.
 
The question asks the year in which the earlier version occurred with a natural language question. To increase linguistic diversity, the prompt is randomly sampled from a curated set of paraphrase templates, which rephrase the temporal query in varied ways. All questions are phrased to elicit a precise, single-year answer. This approach produces a collection of temporally grounded QA pairs designed to evaluate whether models can recall or infer the timing of recurring events. By explicitly requiring models to resolve cross-references between related historical records, the task provides a targeted test of temporal retrieval and factual consistency.

\section{Experiments}
To assess the quality and difficulty of our temporal reasoning benchmarks, we randomly sampled up to 10,000 test examples from each task per language. For languages or benchmarks with fewer examples, we used the entire available set. This leads to a test data with  535843 samples across all languages. 
We evaluated these samples using a suite of multilingual LLMs with diverse architectural and parameter configurations: GPT4o-1120, Mistral-7B-Instruct-v0.2, Gemma-2-9B-IT, Llama-3-8B-Instruct, and Qwen3-8B.
This selection includes both closed and open models, enabling a broad assessment of capabilities across scales and training paradigms.

Our benchmark tasks are designed to probe two complementary aspects of temporal reasoning: (1) Historical factual knowledge, e.g., recognizing when events occurred, even without temporal cues. (2) Temporal inference and reasoning over distant or multi-event sequences, requiring models to understand event ordering and long-range temporal relationships.

Importantly, we evaluate these models in a zero-shot setting without fine-tuning. While fine-tuning can improve reasoning patterns, it cannot substitute for the breadth and accuracy of factual historical knowledge, which must be present in the model's pretraining. This setup ensures that our benchmark serves as a faithful diagnostic of intrinsic temporal reasoning ability, rather than post-hoc adaptation. Prompts used are mentioned in Appendix~\ref{app:zeroShotPrompts}.

\begin{table*}[ht]
\centering
\scriptsize
\begin{tabular}{|l|l|c|c|c|c|c|c|c|c|c|c|}
\hline
\multicolumn{2}{|c|}{Task} & English & Bengali & German & French & Indonesian & Hindi & Italian & Portuguese & Russian & Spanish \\
\hline \hline
\multicolumn{2}{|c|}{FactQA} & 0.488 & 0.489 & 0.433 & 0.472 & 0.603 & 0.525 & 0.366 & 0.659 & 0.581 & 0.589 \\
\multicolumn{2}{|c|}{DurationQA} & 0.318 & 0.805 & 0.886 & 0.791 & 0.893 & 0.815 & 0.889 & 0.875 & 0.906 & 0.826 \\
\hline
\multirow{7}{*}{\rotatebox{90}{RelationQA}}&Compare Duration & 0.732 & 0.557 & 0.777 & 0.615 & 0.771 & 0.737 & 0.826 & 0.739 &  & 0.578 \\
&Duration Diff & 0.138 & 0.065 & 0.302 & 0.094 & 0.327 & 0.243 & 0.196 & 0.435 & & 0.148 \\
&Gap Between & 0.131 & 0.067 & 0.255 & 0.068 & 0.356 & 0.243 &  &  &  &  \\
&Inclusion & 0.707 & 0.083 & 0.806 & 0.388 & 0.863 & 0.585 &  &  &  &  \\
&Order Span End & 0.600 & 0.582 & 0.586 & 0.548 & 0.847 & 0.715 &  &  &  & 0.375 \\
&Order Span Start & 0.841 & 0.790 & 0.888 & 0.723 & 0.913 & 0.848 &  &  &  & 0.844 \\
&Overlap & 0.108 & 0.060 & 0.295 & 0.100 & 0.192 & 0.210 & 0.200 & 0.417 &  & 0.167 \\
\hline
\multicolumn{2}{|c|}{CountQA} & 0.383 & 0.319 & 0.382 & 0.384 & 0.383 & 0.378 & 0.382 & 0.456 & 0.408 & 0.461 \\
\multicolumn{2}{|c|}{SequenceQA} & 0.541 & 0.522 & 0.518 & 0.500 & 0.549 & 0.545 & 0.558 & 0.647 & 0.638 & 0.555 \\
\multicolumn{2}{|c|}{RecurrenceQA} & 0.825 & 0.890 &  &  & 0.858 & 0.952 &  &  &  &  \\
\hline
\end{tabular}
\caption{GPT4o performance across languages on temporal reasoning tasks in our proposed \task{} benchmark. We don't report results for cells with test sample size less than 30.}
\label{tab:gpt4o-multilingual-temporal}
\end{table*}

\begin{table*}[ht]
\centering
\scriptsize
\begin{tabular}{|l|c|c|c|c|c|c|c|c|c|c|}
\hline
 & English & Bengali & German & French & Indonesian & Hindi & Italian & Portuguese & Russian & Spanish \\
\hline \hline
CountQA - Century        & 0.486 & 0.474 & 0.492 & 0.491 & 0.526 & 0.536 & 0.460 & 0.543 & 0.541 & 0.613 \\
CountQA - Between\_events & 0.207 & 0.125 & 0.192 & 0.191 & 0.186 & 0.147 & 0.183 & 0.204 & 0.202 & 0.166 \\
CountQA - Between\_dates  & 0.457 & 0.356 & 0.459 & 0.469 & 0.449 & 0.444 & 0.511 & 0.624 & 0.472 & 0.588 \\
\hline
SequenceQA - Verify      & 0.556 & 0.511 & 0.577 & 0.530 & 0.532 & 0.527 & 0.576 & 0.619 & 0.629 & 0.547 \\
SequenceQA - MCQ         & 0.659 & 0.650 & 0.614 & 0.598 & 0.676 & 0.668 & 0.688 & 0.759 & 0.764 & 0.671 \\
SequenceQA - Arrange     & 0.403 & 0.401 & 0.367 & 0.371 & 0.444 & 0.436 & 0.407 & 0.567 & 0.523 & 0.449 \\
\hline
\end{tabular}
\caption{Detailed performance of GPT4o across languages for CountQA and SequenceQA tasks in our proposed \task{} benchmark.}
\label{tab:qa_type_gpt_results}
\end{table*}

\begin{table*}[ht]
\centering
\scriptsize
\begin{tabular}{|l|l|c|c|c|c|c|c|c|c|c|c|}
\hline
\multicolumn{2}{|c|}{Task} & English & Bengali & German & French & Indonesian & Hindi & Italian & Portuguese & Russian & Spanish \\
\hline \hline
\multicolumn{2}{|c|}{FactQA} & 0.26&0.08&0.22&0.20&0.21&0.10&0.21&0.25&0.23&0.26 \\
\multicolumn{2}{|c|}{DurationQA} & 0.24&0.65&0.68&0.62&0.67&0.63&0.66&0.65&0.66&0.64 \\
\hline
\multirow{7}{*}{\rotatebox{90}{RelationQA}}&Compare Duration & 0.36&0.12&0.40&0.33&0.35&0.30&0.39&0.37&&0.34 \\
&Duration Diff&0.10&0.01&0.14&0.08&0.13&0.08&0.11&0.14&&0.09  \\  
&Gap Between & 0.10&0.01&0.10&0.05&0.12&0.08&&&&  \\  
&Inclusion & 0.40&0.05&0.50&0.30&0.58&0.45&&&&  \\  
&Order Span End & 0.45&0.40&0.48&0.43&0.60&0.50&&&&0.35  \\  
&Order Span Start & 0.60&0.55&0.70&0.60&0.78&0.68&&&&0.68  \\  
&Overlap & 0.08&0.02&0.10&0.06&0.08&0.07&0.07&0.10&&0.08 \\
\hline
\multicolumn{2}{|c|}{CountQA} & 0.29&0.08&0.28&0.24&0.18&0.02&0.27&0.25&0.27&0.26 \\
\multicolumn{2}{|c|}{SequenceQA} & 0.17&0.03&0.17&0.11&0.04&0.10&0.14&0.10&0.16&0.14  \\  \multicolumn{2}{|c|}{RecurrenceQA} & 0.38&0.02&&&0.43&0.02&&&&\\
\hline
\end{tabular}
\caption{Mistral-7B-Instruct performance across languages on temporal reasoning tasks in our proposed \task{} benchmark. We don't report results for cells with test sample size less than 30.}
\label{tab:Mistral-7B-Instruct-multilingual-temporal}
\end{table*}

\begin{table*}[ht]
\centering
\scriptsize
\begin{tabular}{|l|l|c|c|c|c|c|c|c|c|c|c|}
\hline
\multicolumn{2}{|c|}{Task} & English & Bengali & German & French & Indonesian & Hindi & Italian & Portuguese & Russian & Spanish \\
\hline \hline
\multicolumn{2}{|c|}{FactQA} & 0.12&0.10&0.08&0.09&0.11&0.13&0.10&0.14&0.12&0.13\\
\multicolumn{2}{|c|}{DurationQA} & 0.10&0.30&0.35&0.30&0.35&0.32&0.33&0.32&0.34&0.32 \\
\hline
\multirow{7}{*}{\rotatebox{90}{RelationQA}}&Compare Duration & 0.25&0.18&0.27&0.20&0.26&0.28&0.30&0.27&&0.21  \\
&Duration Diff& 0.05&0.02&0.06&0.03&0.07&0.06&0.04&0.06&&0.04\\
&Gap Between & 0.05&0.01&0.04&0.02&0.07&0.05&&&&\\
&Inclusion & 0.25&0.03&0.30&0.18&0.32&0.26&&&&\\
&Order Span End & 0.30&0.28&0.32&0.27&0.36&0.35&&&&0.24 \\
&Order Span Start & 0.45&0.40&0.50&0.42&0.53&0.50&&&&0.50  \\
&Overlap & 0.04&0.01&0.05&0.03&0.04&0.05&0.04&0.07&&0.04 \\
\hline
\multicolumn{2}{|c|}{CountQA} &0.05&0.08&0.07&0.09&0.05&0.05&0.09&0.10&0.11&0.08  \\
\multicolumn{2}{|c|}{SequenceQA} & 0.10&0.10&0.14&0.14&0.17&0.17&0.14&0.17&0.15&0.10\\
\multicolumn{2}{|c|}{RecurrenceQA} &0.65&0.49&&&0.52&0.87&&&& \\
\hline
\end{tabular}
\caption{LLaMA-3-8B-Instruct performance across languages on temporal reasoning tasks in our proposed \task{} benchmark. We don't report results for cells with test sample size less than 30.}
\label{tab:LLaMA-3-8B-Instruct-multilingual-temporal}
\end{table*}

\begin{table*}[ht]
\centering
\scriptsize
\begin{tabular}{|l|l|c|c|c|c|c|c|c|c|c|c|}
\hline
\multicolumn{2}{|c|}{Task} & English & Bengali & German & French & Indonesian & Hindi & Italian & Portuguese & Russian & Spanish \\
\hline \hline
\multicolumn{2}{|c|}{FactQA} & 0.41&0.30&0.38&0.39&0.40&0.36&0.35&0.42&0.41&0.40 \\
\multicolumn{2}{|c|}{DurationQA} & 0.60&0.75&0.79&0.76&0.81&0.73&0.78&0.79&0.82&0.78\\
\hline
\multirow{7}{*}{\rotatebox{90}{RelationQA}}&Compare Duration & 0.62&0.45&0.65&0.59&0.64&0.61&0.67&0.64&&0.63 \\
&Duration Diff& 0.20&0.08&0.22&0.18&0.24&0.22&0.20&0.27&&0.21\\
&Gap Between & 0.18&0.07&0.20&0.17&0.25&0.21&&&&  \\
&Inclusion & 0.60&0.15&0.70&0.45&0.78&0.62&&&& \\
&Order Span End & 0.65&0.55&0.70&0.60&0.80&0.68&&&&0.68\\
&Order Span Start & 0.78&0.70&0.85&0.75&0.90&0.82&&&&0.85 \\
&Overlap & 0.15&0.05&0.18&0.12&0.16&0.14&0.13&0.20&&0.15  \\
\hline
\multicolumn{2}{|c|}{CountQA} & 0.28&0.21&0.28&0.27&0.26&0.24&0.32&0.30&0.29&0.29 \\
\multicolumn{2}{|c|}{SequenceQA} & 0.35&0.26&0.34&0.34&0.31&0.34&0.31&0.39&0.37&0.32 \\
\multicolumn{2}{|c|}{RecurrenceQA} & 0.77&0.71&&&0.80&0.88&&&&   \\
\hline
\end{tabular}
\caption{Gemma-2-9B-IT performance across languages on temporal reasoning tasks in our proposed \task{} benchmark. We don't report results for cells with test sample size less than 30.}
\label{tab:Gemma-2-9B-IT-multilingual-temporal}
\end{table*}

\begin{table*}[ht]
\centering
\scriptsize
\begin{tabular}{|l|l|c|c|c|c|c|c|c|c|c|c|}
\hline
\multicolumn{2}{|c|}{Task} & English & Bengali & German & French & Indonesian & Hindi & Italian & Portuguese & Russian & Spanish \\
\hline \hline
\multicolumn{2}{|c|}{FactQA} & 0.35&0.15&0.31&0.32&0.33&0.24&0.30&0.36&0.35&0.37 \\
\multicolumn{2}{|c|}{DurationQA} & 0.55&0.70&0.72&0.71&0.74&0.68&0.73&0.74&0.76&0.72\\
\hline
\multirow{7}{*}{\rotatebox{90}{RelationQA}}&Compare Duration & 0.55&0.30&0.50&0.48&0.51&0.45&0.53&0.52&&0.54  \\
&Duration Diff& 0.16&0.04&0.18&0.15&0.19&0.16&0.17&0.23&&0.20  \\
&Gap Between & 0.15&0.03&0.15&0.12&0.18&0.14&&&&\\
&Inclusion & 0.55&0.10&0.60&0.42&0.70&0.50&&&&  \\
&Order Span End & 0.58&0.47&0.62&0.55&0.73&0.60&&&&0.64\\
&Order Span Start & 0.73&0.65&0.80&0.70&0.88&0.75&&&&0.80  \\
&Overlap & 0.12&0.03&0.14&0.10&0.13&0.11&0.11&0.16&&0.13 \\
\hline
\multicolumn{2}{|c|}{CountQA} & 0.27&0.06&0.22&0.26&0.20&0.10&0.20&0.21&0.26&0.27  \\
\multicolumn{2}{|c|}{SequenceQA} & 0.28&0.09&0.22&0.23&0.22&0.11&0.25&0.22&0.29&0.25 \\
\multicolumn{2}{|c|}{RecurrenceQA} &0.37&0.22&&&0.22&0.35&&&&  \\
\hline
\end{tabular}
\caption{Qwen3-8B performance across languages on temporal reasoning tasks in our proposed \task{} benchmark. We don't report results for cells with test sample size less than 30.}
\label{tab:Qwen3-8B-multilingual-temporal}
\end{table*}

\section{Results}
We report exact match accuracy for all the 6 tasks across languages. Results for GPT4o are reported in Table~\ref{tab:gpt4o-multilingual-temporal}. Similarly, results for Mistral-7B-Instruct, LLaMA-3-8B-Instruct, Gemma-2-9B-IT, and Qwen3-8B are reported in Tables~\ref{tab:Mistral-7B-Instruct-multilingual-temporal},~\ref{tab:LLaMA-3-8B-Instruct-multilingual-temporal},~\ref{tab:Gemma-2-9B-IT-multilingual-temporal} and~\ref{tab:Qwen3-8B-multilingual-temporal} respectively.

\subsection{Performance by Task Type}
\begin{figure}
    \centering
    \includegraphics[width=0.8\columnwidth]{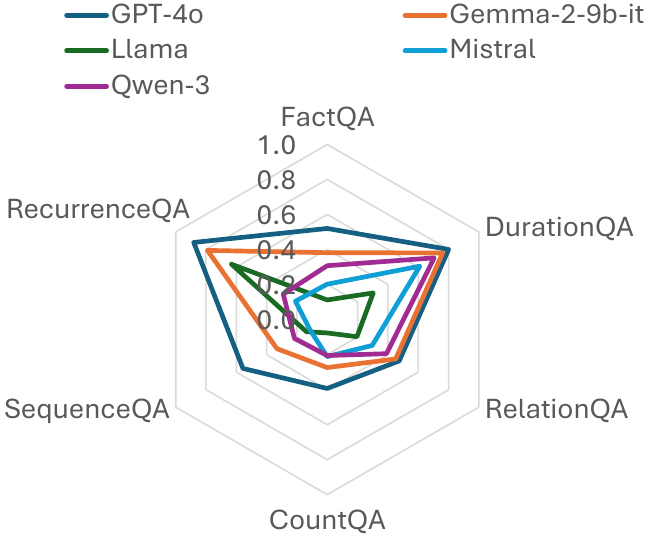}
    \caption{Performance by Question Type}
    \label{fig:perfByTask}
\end{figure}

\begin{figure}
    \centering
    \includegraphics[width=\columnwidth]{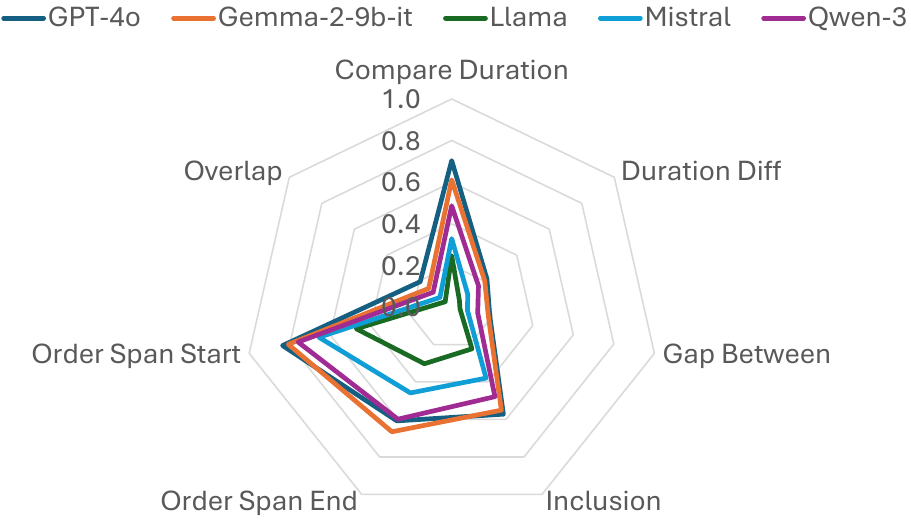}
    \caption{Performance by Question Type for sub-tasks within RelationQA.}
    \label{fig:perfByTask-RelationQA}
\end{figure}

We plot the accuracy for each model by task type in Figs.~\ref{fig:perfByTask} and~\ref{fig:perfByTask-RelationQA}. Fig.~\ref{fig:perfByTask-RelationQA} specifically focuses on sub-tasks within RelationQA. We observe significant variation in model performance across different temporal reasoning tasks. The lowest accuracy across all models is consistently seen in the \textit{Gap Between}, \textit{Overlap}, and \textit{Duration Difference} tasks within the RelationQA benchmark. These tasks demand a multi-step reasoning process that combines temporal information extraction with numerical inference. Specifically, models must correctly (1) identify the start and end dates of two separate event spans, (2) parse these dates into structured formats, and (3) compute the corresponding difference (e.g., number of overlapping days, total gap between spans, or absolute duration differences).

This multi-hop requirement introduces substantial error propagation. Even a small mistake in identifying one of the event dates can cascade into an incorrect final answer. By contrast, tasks like \textit{Order Span Start} show comparatively better performance, possibly because determining which event started earlier can sometimes be resolved using partial temporal cues, even if both spans are not fully parsed.

Sequence ordering tasks also demonstrate moderate performance, suggesting that models are more comfortable with qualitative temporal relationships (e.g., before/after) than quantitative reasoning over time intervals. Overall, these results highlight the challenges models face when asked to internalize structured time and perform arithmetic or set-based reasoning over date spans.

Table \ref{tab:qa_type_gpt_results} lists the sub-task results for CountQA and SequenceQA for GPT4o.  Within the CountQA tasks, the \textit{Between\_events} setting shows the lowest performance. This is expected, as the task implicitly combines three subtasks: (i) identifying the event dates, (ii) sorting the events in temporal order, and (iii) reasoning about how many fall between two given events. Each of these steps introduces potential sources of error, leading to compounding difficulty. In contrast, the \textit{Between\_dates} variant is relatively easier, since it only requires finding the event dates and counting how many lie within a fixed range, without the added complexity of event-to-event comparisons. The \textit{Century} type is somewhat easier still, as it reduces to recognizing the century of given events, which often boils down to identifying whether an event happened in the previous edition or within a well-bounded historical period.  

For the SequenceQA tasks, we also see a natural ordering of difficulty. The \textit{MCQ} setting achieves the highest performance, as the model only needs to identify the correct sequence from a small number of options, which effectively turns the task into a recognition problem. The \textit{Verify} variant is more challenging, requiring the model to judge the correctness of an event sequence in a binary True/False format. Finally, the \textit{Arrange} type is the most difficult, since it demands constructing the chronological order from scratch by reasoning over multiple events, which amplifies the complexity of temporal ordering. Overall, these trends highlight that performance is strongly correlated with the number of reasoning steps involved: tasks requiring multi-step temporal reasoning (e.g., Between\_events, Arrange) consistently exhibit lower scores compared to those that reduce to direct recognition or simpler range checks.

\subsection{Performance by Language}

\begin{figure}[!t]
    \centering
    \includegraphics[width=0.8\columnwidth]{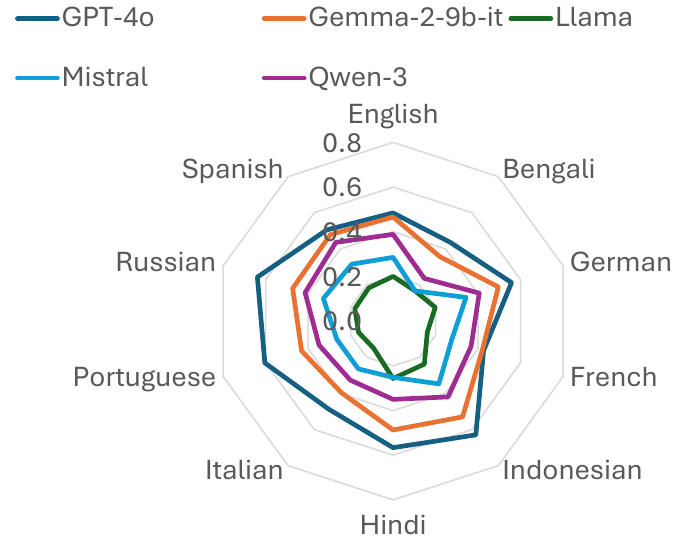}
    \caption{Performance by Language}
    \label{fig:perfByLang}
\end{figure}

We plot the accuracy for each model by language in Figs.~\ref{fig:perfByLang}. Interestingly, despite being relatively low-resource in terms of Wikipedia coverage compared to English or German, Hindi and Indonesian consistently yield the highest average model performance across all benchmark tasks. 
These findings underscore that model performance is not solely a function of language scale but also of how temporally structured information is represented in that language's Wikipedia and how well the model captures its grammatical and formatting conventions.


\subsection{Comparison with GPT4o and Across Smaller Models}

When compared against GPT4o, the four smaller models: LLaMA-3-8B-Instruct, Mistral-7B-Instruct, Gemma-2-9B-IT, and Qwen-3-8B show substantially lower accuracies across almost all tasks and languages. Nonetheless, clear differences emerge in their relative capabilities.
 
Among the smaller models, Gemma-2-9B-IT consistently achieves the highest accuracies across most tasks, often approaching GPT4o performance on structured reasoning problems such as DurationQA, Compare Duration, and Order Span Start. Qwen-3-8B ranks second, showing robust but slightly lower performance than Gemma, particularly on arithmetic-heavy tasks. Mistral-7B-Instruct demonstrates moderate accuracy, performing best in qualitative ordering tasks (e.g., Order Span Start/End) but faltering in numerical reasoning. LLaMA-3-8B-Instruct is the weakest overall, with very low performance on tasks that require multi-step reasoning (Duration Difference, Gap Between, and Overlap). This ranking establishes a clear hierarchy: Gemma $>$ Qwen $>$ Mistral $>$ LLaMA.

Unlike GPT4o, which demonstrates relatively stable multilingual performance (especially in high-resource languages such as Spanish, Portuguese, and Russian), the smaller models exhibit sharper language-level drops. LLaMA struggles particularly in Bengali and Hindi across almost all tasks, often falling below 0.1 accuracy in complex tasks such as Gap Between and Overlap. Mistral shows higher variance across languages, with performance collapsing in Bengali and Hindi, though it recovers somewhat in Romance languages. Gemma and Qwen are more balanced across languages, but still show weaker results in low-resource languages compared to GPT4o.

\paragraph{Task-level Trends.} 
The task-level trends of smaller models broadly mirror those observed for GPT4o:
\begin{itemize}
    \item \textit{Gap Between}, \textit{Overlap}, and \textit{Duration Difference} remain the hardest tasks, with all smaller models performing even worse than GPT4o (often below 0.2). This highlights that fine-grained interval arithmetic is a general weakness, exacerbated in smaller models.
    \item Ordering tasks (\textit{Order Span Start/End}) are comparatively easier. Gemma in particular achieves $>0.8$ in several languages, nearly matching GPT4o, while LLaMA lags significantly.
    \item \textit{DurationQA} shows the largest gap between GPT4o and the smaller models: while GPT4o averages $\sim 0.80$, Gemma and Qwen reach only $0.70$--$0.78$, with Mistral and LLaMA trailing far behind.
    \item \textit{Inclusion} is another area where Gemma and Qwen approach GPT4o, with accuracies $>0.7$ in many Romance languages, while LLaMA and Mistral remain far weaker.
    \item \textit{CountQA} and \textit{SequenceQA} are uniformly low across all models, but GPT4o still outperforms the smaller models by a wide margin.
\end{itemize}

Despite these differences, the overall task difficulty ranking remains consistent with GPT4o: ordering and qualitative reasoning are easier, while numerical reasoning over temporal spans remains the most challenging.


\section{Conclusion}
In this work, we introduce a large-scale multilingual dataset of historical events, \data{}, automatically extracted from Wikipedia timelines and article infoboxes. We highlight how the scale and structure of extracted events vary significantly across languages, with English contributing the largest volume of data. Building on this event database, we construct a comprehensive suite of temporal reasoning tasks within a benchmark called \task{}, that evaluate a model’s ability to understand and reason about time in the context of real-world historical events. We also establish baseline performance metrics using several state-of-the-art multilingual language models. Our findings reveal substantial variation in model performance across both tasks and languages, underscoring the diverse linguistic and temporal reasoning challenges posed by multilingual historical data.

\section{Limitations}
While our work presents a significant advancement in multilingual temporal reasoning over historical events, several limitations remain that warrant future exploration:

\begin{itemize} \item \textbf{Event Coverage Bias:} Our event database is constructed primarily from Wikipedia infoboxes and On This Day pages, which tend to emphasize Western-centric, encyclopedic, and notable events. This introduces a coverage bias, potentially underrepresenting events from marginalized regions, indigenous cultures, and non-mainstream historical narratives.

\item \textbf{Synthetic Question Generation:} The QA datasets rely on automatically generated questions using instruction-tuned models. While this enables scalability, it may introduce artifacts such as unnatural phrasing, limited diversity, or subtle biases in question formulation. Human-authored questions could improve linguistic richness and realism.

\item \textbf{Limited Evaluation Scope:} Our baseline evaluations focus on zero-shot setting using a small set of models. We do not explore few-shot learning, retrieval augmented generation (RAG), or more advanced prompting strategies, which may yield better performance and deeper insights into model capabilities.

\item \textbf{Temporal Reasoning Beyond Factual Recall:} While our benchmark includes tasks that go beyond simple date retrieval, many questions still rely on factual recall rather than complex temporal inference involving causality, counterfactuals, or narrative coherence. Expanding the benchmark to include such dimensions would further enrich the evaluation landscape.

\item \textbf{Lack of Human Evaluation:} We rely on automatic metrics and answer keys derived from structured data. Human evaluation of model responses, especially for open-ended or comparative tasks, could provide more nuanced insights into reasoning quality and answer plausibility.
\end{itemize}


\section{Ethics Statement}
This work involves the construction of a large-scale multilingual dataset of historical events and the development of temporal question answering benchmarks. All data used in this study are derived from publicly available sources, primarily Wikipedia, which is licensed under Creative Commons Attribution-ShareAlike (CC BY-SA). We respect the licensing terms and ensure that our dataset does not include any proprietary or personally identifiable information.

The automatic extraction and generation processes rely on large language models (LLMs), which may introduce biases or inaccuracies. We acknowledge that Wikipedia content itself may reflect cultural, geographic, and editorial biases, which can propagate into our dataset. We have taken steps to mitigate these effects by including diverse languages and domains, but further work is needed to ensure equitable representation across cultures and historical perspectives.

Our benchmark tasks are designed for research purposes only and should not be used in high-stakes decision-making without human oversight. We caution against deploying temporal reasoning models trained on this dataset in sensitive applications such as historical education, policy analysis, or legal contexts without rigorous validation.

Finally, we commit to releasing our dataset and benchmarks under an open license to support transparency, reproducibility, and community-driven improvements. We encourage responsible use and welcome feedback to improve the fairness, inclusivity, and accuracy of our resources.




\bibliography{references}

\appendix
\noindent{\Large \textbf{Overview of Appendices}}
\begin{itemize}
    \item Appendix~\ref{app:detailedDatasetAnalysis}: Detailed Dataset Analysis.
\item Appendix~\ref{appendix:event_extraction_prompt}: Prompt for Event Extraction.
\item Appendix~\ref{appendix:fact_qa_prompt}: Prompt for FactQA question generation.
\item Appendix~\ref{app:durationQA}: Prompts for DurationQA questions.
\item Appendix~\ref{app:zeroShotPrompts}: Prompts for zero-shot evaluation.
\item Appendix~\ref{app:examples}: More examples of English Question Answer Pairs from \task{}.
\end{itemize}

\section{Detailed Dataset Analysis}
\label{app:detailedDatasetAnalysis}
\begin{figure}
    \centering
    \includegraphics[width=\linewidth]{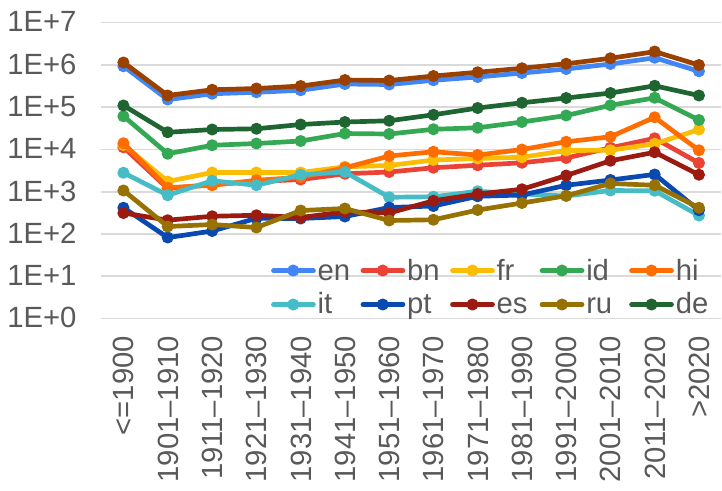}
    \caption{Time-wise Distribution of Events across all languages in \data{}.}
    \label{fig:eventDistributionAllLangs}
\end{figure}

Fig.~\ref{fig:eventDistributionAllLangs} shows time-wise distribution of events across all languages in \data{}. Data across all languages seems to follow the same trend with two notable exceptions: (1) for Italian and Russian, there is a drop from 1940s to 1950s, and (2) for French, there have been a significantly higher proportion of events published in the 2020s (until mid 2025).

\begin{figure*}
    \centering
    \includegraphics[width=\linewidth]{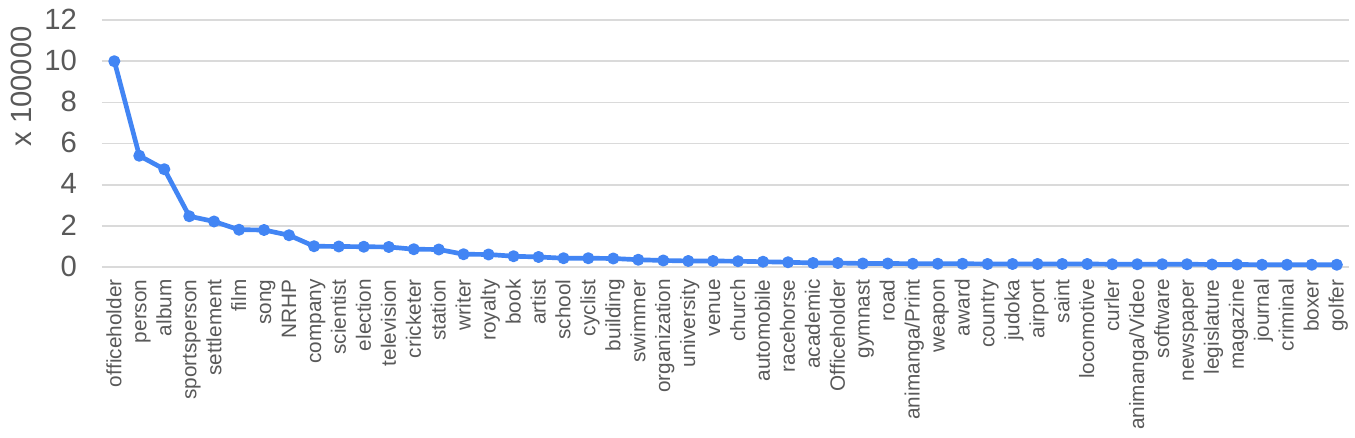}
    \caption{Entity (i.e. Infobox) type Distribution of English Events in \data{}. We show most frequent 50 infobox types for events from other languages in Table~\ref{tab:infoboxTypeDistribution-all}.}
    \label{fig:infoboxTypeDistribution-en}
\end{figure*}

Further, we also how the distribution of English events across various entity (i.e., infobox) types in Fig.~\ref{fig:infoboxTypeDistribution-en}. The distribution of English events across infobox types highlights a strong focus on biographical and cultural entities. ``Officeholder'' leads by a wide margin with over 1 million entries, followed by ``person'', ``album'', and ``sportsperson'', reflecting the prominence of political, individual, and entertainment-related content. Cultural artifacts like films, songs, and books also feature heavily, indicating rich documentation in media domains. Scientific and academic categories such as scientist, university, and academic are present but less dominant. Overall, the data suggests a strong bias toward public figures and popular culture in English-language event documentation.

\begin{table*}[!t]
    \scriptsize
    \begin{tabular}{|l|p{0.9\textwidth}|}
\hline
Lang&Most frequent 50 Infobox Types\\
\hline
\hline
bn&scientist, officeholder, Officeholder, royalty, person, MP, university, company, sportsperson, monarch, organization, station, election, country, book, building, television, legislature, album, Politician, cricketer, writer, spaceflight, holiday, President, software, venue, school, road, airport, philosopher, artist, award, saint, website, song, University, animanga/Video, stadium, economist, School, language, OS, bridge, magazine, president, NRHP, athlete, Country, swimmer\\
\hline
fr&Footballeur, Biographie2, Navire, Monument, Cycliste, Rugbyman, Politicien, Livre, Société, Sportif, Boxeur, Château, Gare, Automobile, Cheval, Saint, Athlète, Biographie, Artiste, Écrivain, Gouvernement, Cérémonie, Organisation2, Handballeur, Pièce, Assemblée, Golfeur, Art, Attentat, Stade, Pont, Route/Base, Événement, Pilote, Criminel, Arme, Musée, biographie2, Scientifique, Récompense, Cimetière, Université, Nageur, Eurovision, Blindé, Nouvelle, Logiciel, Massacre, Gratte-ciel, Exposition\\
\hline
id&officeholder, Officeholder, film, album, settlement, television, royalty, song, company, station, Planet, Album, spesies, scientist, sportsperson, diocese, writer, election, Film, automobile, church, President, planet, animanga/Video, animanga/Print, single, actor, Judge, artist, swimmer, Election, country, book, Politician, saint, building, award, Television, monarch, Person, venue, Airline, language, university, stadium, airport, software, weapon, Sekolah, Company\\
\hline
hi&settlement, cricketer, Film, television, royalty, person, book, writer, Officeholder, scientist, film, sportsperson, spaceflight, album, station, election, university, President, building, country, language, philosopher, company, OS, airport, holiday, University, legislature, organization, Person, software, stadium, artist, legislation, Judge, actor, award, venue, website, spaceflight/IP, Weapon, Company, islands, Writer, Politician, dam, airline, school, Settlement\\
\hline
it&nave, particella, nave/Sandbox, conflitto, cimitero, utente/Wikidipendenza, fiume, colore, album, utente, person, isola, podcast, nave/LinkCategoria, nave/Insegne, nave/Bandiera, nave/Background, malattia, lettere/cirillico, isola/Categorie, fazione militare, elemento chimico/mini tavola periodica, elemento chimico/Colore, Draft NFL, Draft NBA, Affaire criminelle\\
\hline
pt&person, animangá/Anime, empresa, Monarca, VG, ator, animangá/OVA, software, company, Clima, Jornal, website, haplogrupo, animangá/Mangá, television, Album, Rally, language, country, OS, carruagem, weather, atentado, University, automóvel, IFs, Biografia2, film, Weather, Sudão-Estados, mineral, casino, álbum, Société, IYPT, Company, spaceflight, settlement, eleição, Computer, Book, illustrator, Treaty, Sintetizador, Schiff/Basis, Peça, Localidade, Artefato, shopping, organization\\
\hline
es&company, Eurovision, NRHP, officeholder, animanga/Video, film, animanga/Print, automobile, Company, Eurovisión, television, software, person, scientist, NYCS, single, organization, VG, artist, animanga/Header, book, writer, building, filesystem, website, stadium, holiday, Organization, journal, event, animanga/Game, musical, University, Airline, boxer, flag, Software, Election, animanga/Other, OS, zoo, computer, Airport, opera, Book, wildfire, web, airport, Theatre, color\\
\hline
ru&animanga/Manga, software, animanga/Anime, Software, weapon, company, animanga/OVA, person, Company, planet, book, OS, referendum, animanga/Game, animanga/Film, animanga/Other, animanga/Novel, Website, wildfire, Skyscraper, animanga/Header, animanga/Movie, Weather, filesystem, animanga/Drama, Person, Indy500, biodatabase, NRHP, software, writer, Chinese, gene, Website, Software, OS, Disease, Company, winter storm small, terrorist attack, station, sports league, ship image, ship begin, rune, road/hide/tourist, road/hide/states, road/hide/ruralmuni\\
\hline
de&Chartplatzierungen, Leichtathlet, Unternehmen, Film, Band, Schiff, Musikalbum, Schiff/Basis, Tennisspieler, Tennisturnierjahrgang, Eishockeyspieler, Fernsehsendung, Handballer, Radsportler, Fußballsaison, Stadion, Medaillen, Schienenfahrzeug, Biathlet, Publikation, Song, Asteroid, PKW-Modell, Fußballklub, Skilangläufer, Galaxie, Basketballspieler, Verwaltungseinheit, Fluss, Flugzeug, Skispringer, Burg, Volleyballspieler, Schutzgebiet, Brücke, Gesetz, Schule, Chemikalie, Hochschule, Schwimmer, Triathlet, Schiffsklasse/Basis, Fußball-Pokalsaison, Organisation, Boxer, Bahnhof, Episode, Ort, Bauwerk, Eiskunstläufer\\
\hline
    \end{tabular}
    \caption{Most frequent 50 infobox types for events from other languages in \data{}.}
    \label{tab:infoboxTypeDistribution-all}
\end{table*}

Table~\ref{tab:countryList} presents a list of most frequent 50 countries related to the events across languages. English (en) shows a strong Western and Anglophone focus, with the United States, United Kingdom, and Canada leading. In contrast, languages like Bengali (bn) and Hindi (hi) emphasize South Asian countries such as India, Bangladesh, and Pakistan, reflecting regional relevance. European languages like French (fr), Italian (it), and German (de) include a broader mix of European and global countries, while Indonesian (id) and Portuguese (pt) editions show strong representation from Southeast Asia and Latin America, respectively. Overall, the distribution reflects both global commonalities and regional editorial priorities in multilingual event coverage.
\begin{table*}[!t]
    \centering
    \scriptsize
    \begin{tabular}{|l|p{0.9\textwidth}|}
\hline
Lang&Country List\\
\hline
\hline
en&United States, United Kingdom, Canada, Australia, France, India, Japan, Germany, England, Italy, Spain, Russia, China, Brazil, Ireland, Netherlands, Sweden, New Zealand, South Korea, Poland, Mexico, Iran, Scotland, Philippines, Norway, South Africa, Turkey, Argentina, Soviet Union, Ukraine, Pakistan, Belgium, Switzerland, Denmark, Hungary, Czech Republic, Greece, Malaysia, Finland, Indonesia, Austria, Romania, Portugal, Israel, Serbia, Thailand, Nigeria, Global, Chile, Croatia\\
\hline
bn&India, Bangladesh, United States, United Kingdom, Pakistan, England, France, Japan, Germany, China, Australia, British India, Spain, Iran, Nepal, Afghanistan, Sri Lanka, Russia, Global, Israel, Saudi Arabia, Italy, Canada, Turkey, Iraq, Soviet Union, Brazil, Egypt, Netherlands, New Zealand, Indonesia, South Africa, Argentina, Sweden, United Arab Emirates, Ottoman Empire, Syria, South Korea, Singapore, Malaysia, Thailand, Yemen, Poland, Portugal, Belgium, Ireland, Scotland, Switzerland, Ukraine, West Indies\\
\hline
fr&France, United States, United Kingdom, Canada, Italy, Germany, Russia, Spain, Japan, Belgium, Switzerland, South Korea, Ukraine, Netherlands, Brazil, England, Algeria, Australia, China, Soviet Union, Mexico, Morocco, Norway, Portugal, Argentina, Poland, Sweden, Ireland, Austria, Colombia, New Zealand, Denmark, Israel, Turkey, Egypt, Finland, South Africa, Chile, India, Hungary, Peru, Europe, Senegal, Croatia, Cameroon, Greece, Romania, Nigeria, Ivory Coast, Thailand, Democratic Republic of the Congo\\
\hline
id&Indonesia, United States, Japan, South Korea, India, France, Italy, Germany, England, United Kingdom, Russia, Global, China, Malaysia, Australia, Spain, Netherlands, Canada, Soviet Union, Brazil, Philippines, Thailand, Israel, Sweden, Poland, Taiwan, Austria, Switzerland, Hong Kong, Turkey, Denmark, Iran, Ireland, Egypt, Argentina, Singapore, Portugal, Mexico, Hungary, Belgium, Greece, Vietnam, Ukraine, Europe, Dutch East Indies, Saudi Arabia, Serbia, Norway, Romania, United Nations, South Africa, Finland, Vatican\\
\hline
hi&India, Pakistan, United States, Australia, England, Bangladesh, Sri Lanka, South Africa, Nepal, United Kingdom, New Zealand, Ireland, West Indies, Japan, Spain, China, Scotland, France, Canada, Zimbabwe, Germany, United Arab Emirates, Netherlands, Afghanistan, Global, Russia, British India, Thailand, Italy, Oman, Soviet Union, Uganda, Kenya, Iran, Sweden, Papua New Guinea, Britain, Namibia, Brazil, Iraq, Denmark\\
\hline
it&Italy, United Kingdom, United States, Japan, Germany, France, Soviet Union, Russia, Spain, Netherlands, Denmark, Finland, Canada, Austria-Hungary, Norway, Republic of Venice, Australia, Greece, Poland, Argentina, Turkey, China, India, Sweden, Austria, Croatia, Panama, South Korea, Kingdom of the Two Sicilies, Portugal, Bahamas, Ukraine, Philippines, Chile, New Zealand, West Germany, Peru, Dutch Republic, Egypt, Romania, Ottoman Empire, Brazil, Yugoslavia, England, Malta, Thailand, Not specified, Ireland, Taiwan\\
\hline
pt&United States, Brazil, Japan, Portugal, Germany, France, Italy, United Kingdom, South Korea, Russia, China, Australia, Canada, Spain, Israel, Sweden, Singapore, Argentina, Denmark, New Zealand, Venezuela, Soviet Union, Global, Belgium, Mexico, Norway, Austria, Uruguay, Chile, Ireland, Finland, Switzerland, Colombia, Greece, Yugoslavia, Netherlands, Turkey, England, South Africa, Serbia, Peru, Poland, Europe, Bulgaria, Cuba, Scotland, Paraguay, Azerbaijan, Ukraine, Hungary, Jamaica, Malta, Croatia\\
\hline
es&Japan, United States, Spain, Mexico, Brazil, Australia, United Kingdom, France, Germany, Italy, Canada, Argentina, Chile, UNASUR, South Korea, Philippines, Venezuela, Portugal, Peru, Colombia, Thailand, China, Malaysia, Taiwan, Singapore, Global, India, Saudi Arabia, Global except Asia, Indonesia, Southeast Asia, Poland, Ecuador, Panama, Hong Kong, Latin America, Netherlands, South Asia, North America, Cambodia, Austria, Laos, Europe, South Africa, Costa Rica, Dominican Republic, New Zealand, Norway, Russia, Vietnam, El Salvador\\
\hline
ru&United States, Japan, Russia, Germany, Soviet Union, United Kingdom, France, Italy, Canada, Global, Finland, China, Switzerland, Spain, Poland, Ukraine, Empire of Japan, Austria, Sweden, Australia, Belgium, Confederate States of America, Norway, Hungary, South Korea, Philippines, German Empire, Croatia, Czechoslovakia, Netherlands, Taiwan, Mexico, Brazil, Russian Empire, Europe, Austria-Hungary, Yugoslavia, Belarus, Argentina, Denmark, Bulgaria\\
\hline
de&Germany, United States, France, United Kingdom, Austria, Japan, Italy, Switzerland, Russia, Canada, Poland, Spain, Netherlands, Sweden, Australia, China, Norway, Turkey, Brazil, Czech Republic, England, Finland, Denmark, Belgium, Soviet Union, Hungary, South Korea, Romania, India, Ukraine, Peru, Argentina, Mexico, Portugal, Greece, Thailand, Slovakia, New Zealand, South Africa, Global, Croatia, Slovenia, Ireland, Colombia, Scotland, Czechoslovakia, Chile, Uruguay, Serbia, Kazakhstan, Estonia, Belarus\\
\hline

    \end{tabular}
    \caption{List of most frequent 50 countries related to the events across languages in \data{}.}
    \label{tab:countryList}
\end{table*}

\begin{figure}
    \centering
    \includegraphics[width=\columnwidth]{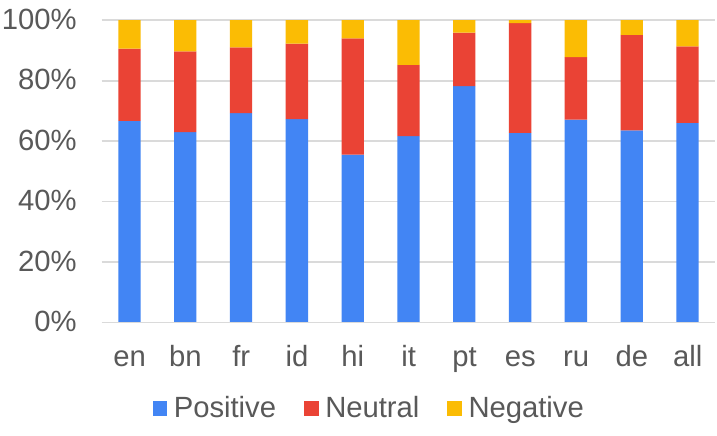}
    \caption{Polarity Distribution of Events across all languages in \data{}}
    \label{fig:polarityDist}
\end{figure}

Fig.~\ref{fig:polarityDist} shows polarity distribution of events across all languages in \data{} dataset. It reveals interesting editorial tendencies. Most languages show a strong preference for positive polarity, with Portuguese (pt) leading at over 78\%, followed by French (fr) and Indonesian (id). Neutral polarity is notably higher in Hindi (hi) and Spanish (es), suggesting a more balanced or factual reporting style. Negative polarity remains relatively low across the board, with Spanish (es) being an outlier at under 1\%, while Italian (it) and Russian (ru) show slightly higher negative sentiment. Overall, the dataset reflects a global inclination toward documenting events with a positive or neutral tone, with limited emphasis on negativity.

Table~\ref{tab:onthisday} shows data statistics per language for events mined from On This Day wikipedia pages.

\begin{table}[!t]
    \centering
    \scriptsize
    \begin{tabular}{|l|c|c|c|}
    \hline
         Lang&birth&death&generic events\\
         \hline
         \hline
bn&6968&5117&6374\\
\hline
en&80252&37248&18941\\
\hline
fr&59387&35644&11979\\
\hline
de&93244&59863&17258\\
\hline
hi&1655&848&2892\\
\hline
id&6081&0&6276\\
\hline
it&0&0&14917\\
\hline
pt&60108&17605&14804\\
\hline
ru&25028&15078&18421\\
\hline
es&68852&28295&24995\\
\hline
    \end{tabular}
    \caption{Data statistics per language for events mined from On This Day wikipedia pages}
    \label{tab:onthisday}
\end{table}

\section{Prompt for Event Extraction} \label{appendix:event_extraction_prompt}
\begin{lstlisting}[caption={Prompt for Event Extraction}, label={lst:eventExtraction}]
Wikipedia has infoboxes. We have extracted the infobox data in a JSON format. Each infobox contains the infobox type along with the metadata. The metadata consists of key-value pairs describing the infobox. Some of the key-value pairs may contain dates and those dates might be related to an event. Given the infobox data, filter out the key-value pairs which have dates. Parse the value to extract the date in a year, month, day format. Additionally, generate an event description which happened on that date. Additionally, identify all possible countries to which the event is associated. Also output a polarity (positive, negative, neutral) for each of the possible countries indicating whether the people from the country would like the event. Consider if the event had any negative impact. Keep in mind that there might be multiple events inside a single infobox. Return the output as a list of JSON where each JSON contains the extracted date and event description. Each JSON should also contain the key-value where the date came from. Do not return anything else other than the list.
\end{lstlisting}

\section{Prompt for FactQA question generation}
\label{appendix:fact_qa_prompt}
\begin{lstlisting}[caption={Prompt for FactQA question generation}, label={lst:factQA}]
You are a helpful assistant that generates natural language questions asking for the **date of a specific event**.

    Context: The event is described as:
    - Title: <article title>
    - Event Description: <event description>
    - Infobox Type: <infobox type>
    - Key: <date key>
    - Language: <Language>
    Your task:
    - Write a fluent, natural-sounding **question** in <Language> that asks for the **date** of the above event.
    - Use the event description and title to precisely identify the event. Avoid vague or generic phrasing.
    - The question must be self-contained, clearly referring to the entity or topic involved.
    - The output must be a **question only**. Do not include answers, metadata, or explanations.
    - If a valid question cannot be formed (e.g., due to incomplete or irrelevant data), return **INVALID**.
    Question:
\end{lstlisting}

\section{Prompts for DurationQA questions}
\label{app:durationQA}
\subsection{Prompt for checking DurationQA question validity} \label{appendix:duration_classify_prompt}
\begin{lstlisting}[caption={Prompt for checking DurationQA question validity}, label={lst:duration_classify_prompt}]
You are a knowledge validation assistant.

For infobox type <infobox type>, determine whether the following two events can be used to define a coherent duration.

- Event 1: '{k1}' - "{d1}"

- Event 2: '{k2}' - "{d2}"

Should these two events be used to form a valid start-end duration?

Reply with 'Yes' or 'No' only.
\end{lstlisting}

\subsection{Prompt for DurationQA question generation}\label{appendix:duration_generation_prompt}

\begin{lstlisting}[caption={Prompt for DurationQA question generation}, label={lst:duration_generation_prompt}]
You are a helpful assistant that writes specific duration-based questions between two well-described events.

Context: This is about <article title> from a <infobox type> infobox.

Here are two events with exact descriptions:

1. <article title> - <start event description>

2. <article title> - <end event description>

Write a natural language specific question about entity (not an answer or a general question) that asks about the time duration between these two events.

Important:

- Refer to each event using its description and title exactly as given above and disambiguate the general information with exact event for the entity.

- Do NOT use generic questions without entity like -> 'How many years passed between the release of the album and the release of the single?'

- The question must not be general without entity, it is a well formed independent question which is clear without needing any external information.

- The output should be a **question only** OR INVALID if no valid question or incomplete question can be formed.

Question:
\end{lstlisting}

\section{Prompts for zero-shot evaluation}
\label{app:zeroShotPrompts}
\begin{lstlisting}[caption={Prompt for FactQA zero-shot inference}, label={lst:factQAInf}]
You are an expert in history with extensive knowledge about past events. Find out the dates when each of the given events occurred and answer the question based on your world knowledge and reasoning. STRICTLY return the answer ONLY in a Json in the format in English:
<output>
{
"answerOut": {'year': year, 'month': month, 'day': day},
"reasoning": reasoning
}
</output>
#question#
\end{lstlisting}

\begin{lstlisting}[caption={Prompt for DurationQA zero-shot inference}, label={lst:DurationQAInf}]
You are an expert in history with extensive knowledge about past events. Find out the dates when each of the given events occurred and answer the question based on your world knowledge and reasoning. STRICTLY return the answer ONLY in a Json in the format in English:
<output>
{
"answerOutStart": {'year': year, 'month': month, 'day': day},
"answerOutEnd": {'year': year, 'month': month, 'day': day},
"reasoning": reasoning
}
</output>

#question#
\end{lstlisting}

\begin{lstlisting}[caption={Prompt for RelationQA Compare Duration zero-shot inference}, label={lst:RelationQACompareDurationInf}]
You are an expert in history with extensive knowledge about past events. Find out the which event span is of longer duration given events occurred and answer the question based on your world knowledge and reasoning. STRICTLY return the answer ONLY in a Json in the format in English:
<output>
{
"answerOut": event span,
"reasoning": reasoning
}
</output>

#question#
\end{lstlisting}

\begin{lstlisting}[caption={Prompt for RelationQA Duration Diff, Gap Between, and Overlap zero-shot inference}, label={lst:RelationQADurationDiffGapBetweenOverlapInf}]
You are an expert in history with extensive knowledge about past events. Find out the dates when each of the given event span started and ended and answer the question based on your world knowledge and reasoning. STRICTLY return the answer ONLY in a Json in the format in English:
<output>
{
"answerOutEventSpan1": start year-start month-start day/end year-end month-end day,
"answerOutEventSpan2": start year-start month-start day/end year-end month-end day,
"reasoning": reasoning
}
</output>

#question#
\end{lstlisting}

\begin{lstlisting}[caption={Prompt for RelationQA Order Span Start zero-shot inference}, label={lst:RelationQAOrderSpanStartInf}]
You are an expert in history with extensive knowledge about past events. Find out the order for which event group has the start date first than the other based on the starting event in each group and answer the question based on your world knowledge and reasoning. We are not talking about duration we are talking about comparison between starting event in each event group. STRICTLY return the answer ONLY in a Json in the format in English:
<output>
{
"answerOut": event group started first,
"reasoning": reasoning
}
</output>

#question#
\end{lstlisting}

\begin{lstlisting}[caption={Prompt for RelationQA Order Span End zero-shot inference}, label={lst:RelationQAOrderSpanEndInf}]
You are an expert in history with extensive knowledge about past events. Find out the order for which event group has the end date later than the other based on the ending event in each group and answer the question based on your world knowledge and reasoning. We are not talking about duration we are talking about comparison between ending event in each event group. STRICTLY return the answer ONLY in a Json in the format in English:
<output>
{
"answerOut": event group ending last,
"reasoning": reasoning
}
</output>

#question#
\end{lstlisting}

\begin{lstlisting}[caption={Prompt for RelationQA Inclusion zero-shot inference}, label={lst:RelationQAInclusionInf}]
You are an expert in history with extensive knowledge about past events. Find out the which event group includes in itself the other event group based on the start and end date of events within each group and answer the question based on your world knowledge and reasoning. We are not talking about comparing the duration of events groups we are talking about inclusion of one event group within other group based on starting and ending of events within each group. STRICTLY return the answer ONLY in a Json in the format in English:
<output>
{
"answerOut": event group or NA,
"reasoning": reasoning
}
</output>

#question#
\end{lstlisting}

\begin{lstlisting}[caption={Prompt for CountQA, SequenceQA and RecurrenceQA zero-shot inference}, label={lst:CountQASequenceQARecurrenceQAInf}]
You are an expert in history with extensive knowledge about past events. Find out the dates when each of the given events occured and answer the question based on your world knowledge and reasoning. Return the answer in a json in the format in English:
<output>
{
"answer": answer,
"reasoning": reasoning
}
</output>

#question#
\end{lstlisting}

\section{More examples of English Question Answer Pairs from \task{}}
\label{app:examples}
Tables~\ref{tab:questions_english_part1},~\ref{tab:questions_english_part2} and~\ref{tab:questions_english_part3} show multiple examples of English questions from our dataset for each question type.

\begin{table*}[ht]
\centering
\scriptsize
\begin{tabular}{|p{0.1\textwidth}|p{0.6\textwidth}|p{0.18\textwidth}|}
\hline
Question Type & Question & Answer \\ 
\hline
 \hline
FactQA & I'm looking for the date of the release of the Goodnight Punpun Omnibus 1. when was it released? & \{'year': '2016', 'month': '03', 'day': '04'\} \\ \hline
FactQA & Impax Asset Management Group plc reported £34.4 billion in assets under management in what year? & \{'year': '2021', 'month': '06', 'day': '30'\} \\ \hline \hline
DurationQA & What was the duration between the birth and death of Scarface John Williams? & start\_date \{'year': '1938', 'month': '10', 'day': '19'\} end\_date \{'year': '1972', 'month': '03', 'day': '04'\} \\ \hline
DurationQA & What was the time duration between the release of the single ``Happy Happy'' and the release of the single ``Fake \& True''? & start\_date \{'year': '2019', 'month': '06', 'day': '12'\} end\_date \{'year': '2019', 'month': '10', 'day': '18'\} \\ \hline \hline
CountQA - century & The following is a list of historical events. Each line includes a description and the title of the article it comes from. (1) Event about `Sean Kingston (album)': The single ``Beautiful Girls'' by Sean Kingston was released. (2) Event about `Week End (X Japan song)': The song ``Week End'' by X Japan was released. (3) Event about `Lucy McEvoy': Lucy McEvoy was nominated for the AFL Women's Rising Star award. (4) Event about `BL 5.4-inch howitzer': The Ordnance BL 5.4-inch howitzer was used in the Second Boer War. (5) Event about `Holten Castenschiold': Holten Castenschiold began his term as the 6th President of the Danish Olympic Committee. (6) Event about `William L. Baird': William Lewis Baird began his term as the 19th Mayor of Lynn, Massachusetts. (7) Event about `Jung Yeon-kyung': Jung Yeon-kyung was born. Count the number of events that occurred during the 19th century. Provide the count. & 2 \\ \hline
CountQA - century & The following is a list of historical events. Each line includes a description and the title of the article it comes from. (1) Event about `Departmental Council of Côte-d'Or': François Sauvadet was elected as President of the Departmental Council of Côte-d'Or. (2) Event about `Constellation Place': Construction of Constellation Place began. (3) Event about `Monroe Gooden': Monroe Washington Gooden was born. (4) Event about `Bradenton Marauders': The Bradenton Marauders won the second half championship. (5) Event about `Koguva': The population of Koguva was recorded as 30.List how many events fall in the 21th century. Provide the count. & 3 \\
\hline
CountQA - between\_events & The following is a list of historical events. Each line includes a description and the title of the article it comes from. (1) Event about `Moisés Paniagua': Moisés Paniagua was born. (2) Event about `Ben Rohrer': Ben Rohrer played for Delhi Daredevils. (3) Event about `Home Video (album)': The single 'Going Going Gone' from the album `Home Video' was released. (4) Event about `Black Like Me (film)': The film ``Black Like Me'' was released. Count the number of events that took place after event (4) but before event (2) Do not include event (4) or event (2) in the count. & 1 \\ \hline
CountQA - between\_events & The following is a list of historical events. Each line includes a description and the title of the article it comes from. (1) Event about `Buckeye Township, Michigan': Buckeye Township, Michigan was established. (2) Event about `The Models (Mongolian TV series)': The television show first aired. (3) Event about `Fossil Bluff': Fossil Bluff Station was established. (4) Event about `Royal Noble Consort Uibin Seong': Ui-bin Seong began her tenure as Royal Noble Consort of the First Senior Rank. (5) Event about `Assault Battalion No. 5 (Rohr)': Assault Battalion Number 5 (Rohr) was disbanded. What is the number of events happening strictly between event (5) and event (2)? Do not include event (5) or event (2) in the count. & 2 \\ \hline
CountQA - between\_dates & The following is a list of historical events. Each line includes a description and the title of the article it comes from. (1) Event about `Wang Dang Doodle': The song ``Wang Dang Doodle'' was recorded. (2) Event about `1995–96 Football League': The 1995–96 Football League Second Division season concluded. (3) Event about `Simon Bollom': Sir Simon Bollom was born. (4) Event about `Bahaa Taher': Bahaa Taher was awarded the Arabic Booker Prize.Count the events that occurred from 1990 to 2100. Provide the total number. Include events from both 1990 and 2100. & 2 \\ \hline
CountQA - between\_dates & The following is a list of historical events. Each line includes a description and the title of the article it comes from. (1) Event about `Franck Belot': Franck Belot was born. (2) Event about `Sorato Anraku': Sorato Anraku won a gold medal in Bouldering at the Innsbruck 2023 IFSC Climbing World Cup. (3) Event about `SV Babelsberg 03': SV Babelsberg 03 was founded. (4) Event about `Barack Obama vs. Mitt Romney (video)': The previous song ``Frank Sinatra vs. Freddie Mercury'' was released. (5) Event about `1919 Ayvalık earthquake': The 1919 Ayvalık earthquake occurred, causing significant destruction and loss of life. (6) Event about `Mananciais do Rio Paraíba do Sul Environmental Protection Area': The Mananciais do Rio Paraíba do Sul Environmental Protection Area was created. (7) Event about `Akshay Venkatesh': Akshay Venkatesh was awarded the Infosys Prize. Within the time span of 1920–2050, how many of these events occurred? Provide the total number. Include events from both 1920 and 2050. & 5 \\ \hline 
\end{tabular}
\caption{Examples of Question-Answers pairs for FactQA, DurationQA and CountQA question types in English}
\label{tab:questions_english_part1}
\end{table*}

\begin{table*}[ht]
\centering
\scriptsize
\begin{tabular}{|p{0.1\textwidth}|p{0.6\textwidth}|p{0.18\textwidth}|}
\hline
Question Type & Question & Answer \\ 
\hline
 \hline
Compare Duration & Which event span is of longer duration: ``Agnes van Ardenne was born'' to ``Agnes van Ardenne started her earlier term as Member of the House of Representatives'', or ``Agnes van Ardenne ended her earlier term as Member of the House of Representatives'' to ``Agnes van Ardenne started her term as Member of the House of Representatives.'' in the context of ``Agnes van Ardenne''? & ``Agnes van Ardenne was born'' to ``Agnes van Ardenne started her earlier term as Member of the House of Representatives.'' \\ \hline
Compare Duration & Which event span is of longer duration: ``Albert Planasdemunt i Gubert began his term as a Member of the Parliament of Catalonia'' to ``Albert Planasdemunt i Gubert began his term as Mayor of Breda''', or ``Albert Planasdemunt i Gubert was born'' to ``Albert Planasdemunt i Gubert passed away.'' in the context of ``Albert Planasdemunt i Gubert''? & ``Albert Planasdemunt i Gubert was born'' to ``Albert Planasdemunt i Gubert passed away.'' \\ \hline \hline
Duration Diff & What is the difference in duration between ``A new session of the Alabama Legislature began'' to ``The next election for the Alabama Senate is scheduled'', and ``Anthony Daniels was elected as House Minority Leader of Alabama'' to ``The last election for the Alabama Senate was held.'' (in days) in the context of ``Alabama Legislature''? & 762 \\ \hline
Duration Diff & What is the difference in duration between ``Adolf von Thadden was born'' to ``Adolf von Thadden began his term as a Member of the Bundesrat.'' and ``Adolf von Thadden ended his term as a Member of the Bundestag'' to ``Adolf von Thadden ended his term as a Member of the Bundesrat.'' (in days) in the context of ``Adolf von Thadden''? & 10280 \\ \hline \hline
Gap Between & What is the gap in days between ``José Luis Soro became the leader of Chunta Aragonesista'' to ``Maru Díaz became the leader of Podemos–Green Alliance in Aragon.'', and ``Tomás Guitarte became the leader of Teruel Existe'' to ``Alberto Izquierdo became the leader of the Aragonese Party.'', in the context of ``2023 Aragonese regional election''? & 1522 \\ \hline
Gap Between & What is the gap in days between ``The January 2007 special session of the 98th Wisconsin Legislature started'' to ``The March 2008 special session of the 98th Wisconsin Legislature started'', and ``The December 2007 special session of the 98th Wisconsin Legislature ended'' to ``The April 2008 special session of the 98th Wisconsin Legislature ended.'', in the context of ``98th Wisconsin Legislature''? & 62 \\ \hline \hline
Inclusion & Which event's time span includes the other: ``June 2009 Extra Session of the 99th Wisconsin Legislature ended'' to ``December 2009 Special Session of the 99th Wisconsin Legislature began'', or ``Election for the 99th Wisconsin Legislature was held'' to ``The term of the 99th Wisconsin Legislature ended.'', in the context of ``99th Wisconsin Legislature''? & ``Election for the 99th Wisconsin Legislature was held'' to ``The term of the 99th Wisconsin Legislature ended.'' \\ \hline
Inclusion & Which event's time span includes the other: ``Ahmad Maslan ended his term as Deputy Minister of Finance'' to ``Ahmad Maslan began his term as State Deputy Chairman of the United Malays National Organisation of Johor'', or ``Ahmad Maslan ended his term as Deputy Minister of International Trade and Industry'' to ``Ahmad Maslan began his term as Secretary-General of Barisan Nasional.'', in the context of ``Ahmad Maslan''? & ``Ahmad Maslan ended his term as Deputy Minister of Finance'' to ``Ahmad Maslan began his term as State Deputy Chairman of the United Malays National Organisation of Johor.'' \\ \hline \hline 
Order Span End & Which event span ended last: ``The album `100 Reasons to Live' by Gareth Emery was released'' to ``The single `Far From Home' by Gareth Emery feat. Gavrielle was released'', or ``The single `Hands' by Gareth Emery \& Alastor feat. London Thor was released'' to ``The single `Save Me' by Gareth Emery was released.'', in the context of ``100 Reasons to Live''? & ``The single `Hands' by Gareth Emery \& Alastor feat. London Thor was released'' to ``The single `Save Me' by Gareth Emery was released''. \\ \hline
Order Span End & Which event span ended last: ``The March 2018 extraordinary session of the 103rd Wisconsin Legislature started'' to ``The 103rd Wisconsin Legislature term ended'', or ``The January 2018 special session of the 103rd Wisconsin Legislature started'' to ``The March 2018 extraordinary session of the 103rd Wisconsin Legislature ended.'', in the context of ``103rd Wisconsin Legislature''? & ``The March 2018 extraordinary session of the 103rd Wisconsin Legislature started'' to ``The 103rd Wisconsin Legislature term ended.'' \\ \hline \hline
Order Span Start & Which event span started earlier: ``The album `100 Reasons to Live' by Gareth Emery was released'' to ``The single `Far From Home' by Gareth Emery feat. Gavrielle was released'', or ``The single `Hands' by Gareth Emery \& Alastor feat. London Thor was released'' to ``The single `Save Me' by Gareth Emery was released.'', in the context of ``100 Reasons to Live''? & ``The single `Hands' by Gareth Emery \& Alastor feat. London Thor was released'' to ``The single `Save Me' by Gareth Emery was released.'' \\ \hline
Order Span Start & Which event span started earlier: ``The March 2018 extraordinary session of the 103rd Wisconsin Legislature started'' to ``The 103rd Wisconsin Legislature term ended'', or ``The January 2018 special session of the 103rd Wisconsin Legislature started'' to ``The March 2018 extraordinary session of the 103rd Wisconsin Legislature ended.'', in the context of ``103rd Wisconsin Legislature''? & ``The January 2018 special session of the 103rd Wisconsin Legislature started'' to ``The March 2018 extraordinary session of the 103rd Wisconsin Legislature ended.'' \\ \hline \hline
Overlap & How many days do ``The single `Hands' by Gareth Emery \& Alastor feat. London Thor was released'' to ``The single `Far From Home' by Gareth Emery feat. Gavrielle was released.'', and ``The album '100 Reasons to Live' by Gareth Emery was released'' to ``The single 'Save Me' by Gareth Emery was released.'' overlap, in the context of ``100 Reasons to Live''? & 29 \\ \hline
Overlap & How many days do ``The 2016 Wisconsin elections were held'' to ``The November 2018 extraordinary session of the 103rd Wisconsin Legislature started.'', and ``The January 2017 special session of the 103rd Wisconsin Legislature ended'' to ``The March 2018 special session of the 103rd Wisconsin Legislature started.'' overlap, in the context of ``103rd Wisconsin Legislature''? & 275 \\ \hline
\end{tabular}
\caption{Examples of Question-Answers pairs for RelationQA question type in English}
\label{tab:questions_english_part2}
\end{table*}

\begin{table*}[ht]
\centering
\scriptsize
\begin{tabular}{|p{0.1\textwidth}|p{0.6\textwidth}|p{0.18\textwidth}|}
\hline
Question type&Question&Answer\\ \hline  \hline
SequenceQA - Verify & Is the following timeline historically valid? Each event includes a description and the title of the article it came from. Answer `True' if the events are in correct chronological order, otherwise `False'. (1) Event about `Mudhoji II of Nagpur': Mudhoji II began his reign as the 6th Raja of Nagpur. (2) Event about `L'Épiphanie': L'Épiphanie was officially constituted as a city. (3) Event about `Dossena': The population of Dossena was recorded as 966. (4) Event about `Next (Indian retailer)': The company profile page of Next Retail India Ltd was archived. (5) Event about `Richard Money': Richard Money began managing Hartlepool United. & FALSE \\ \hline
SequenceQA - Verify & Do the events occur in proper chronological sequence? Each event includes a description and the title of the article it came from.Answer `True' if the events are in correct chronological order, otherwise `False'. (1) Event about `River Vale Skeeters': The River Vale Skeeters hockey team began operations. (2) Event about `Sandrine Hamel': Sandrine Hamel won a bronze medal in Dual banked slalom at the 2021 World Para Snow Sports Championships in Lillehammer. (3) Event about `Dominique Walter': Dominique Walter was born. (4) Event about `Castaic Dam': Castaic Dam was opened. & FALSE \\
\hline\hline
SequenceQA - MCQ & Only one of the sequences below is in the correct order. Can you find it? Each event is described alongside its article title. Choose the correct order using the event descriptions and article titles provided. (A) `Zaza Nadiradze': Zaza Nadiradze was born. | `ND Slovan': Nogometno društvo Slovan was founded. | `Daniele Rimpelli': Daniele Rimpelli represented the Italy national rugby union team. | `Matt Viney': Matt Viney began his term as a Member of the Victorian Legislative Council for Eastern Victoria Region. (B) `ND Slovan': Nogometno društvo Slovan was founded. | `Daniele Rimpelli': Daniele Rimpelli represented the Italy national rugby union team. | `Matt Viney': Matt Viney began his term as a Member of the Victorian Legislative Council for Eastern Victoria Region. | `Zaza Nadiradze': Zaza Nadiradze was born. (C) `ND Slovan': Nogometno društvo Slovan was founded. | `Zaza Nadiradze': Zaza Nadiradze was born. | `Matt Viney': Matt Viney began his term as a Member of the Victorian Legislative Council for Eastern Victoria Region. | `Daniele Rimpelli': Daniele Rimpelli represented the Italy national rugby union team. (D) `Matt Viney': Matt Viney began his term as a Member of the Victorian Legislative Council for Eastern Victoria Region. | `Zaza Nadiradze': Zaza Nadiradze was born. | `Daniele Rimpelli': Daniele Rimpelli represented the Italy national rugby union team. | `ND Slovan': Nogometno društvo Slovan was founded. & C \\ \hline
SequenceQA - MCQ & Select the option that shows the correct chronological order of events. Each event is described alongside its article title.Choose the correct order using the event descriptions and article titles provided. (A) `Darrel Higham': Darrel Higham was born. | `Type 620 tanker': The ship Shengli had its maiden voyage. | `Sian Williams': Sian Williams married Neale Hunt. (B) `Type 620 tanker': The ship Shengli had its maiden voyage. | `Darrel Higham': Darrel Higham was born. | `Sian Williams': Sian Williams married Neale Hunt. (C) `Type 620 tanker': The ship Shengli had its maiden voyage. | `Sian Williams': Sian Williams married Neale Hunt. | `Darrel Higham': Darrel Higham was born. (D) `Darrel Higham': Darrel Higham was born. | `Sian Williams': Sian Williams married Neale Hunt. | `Type 620 tanker': The ship Shengli had its maiden voyage. & A \\ \hline \hline
SequenceQA - Arrange & Sort the events by the time they took place. Each event is accompanied by a description and the title of the article it comes from. Return the correct chronological order by listing the event numbers, like (2) (1) (3). (1) Event about `Paaliaq': Paaliaq was discovered. (2) Event about `Bob Chakales': Bob Chakales was born. (3) Event about `Dark Valley': The film ``Dark Valley'' was released. & (2) (3) (1) \\ \hline
SequenceQA - Arrange & Given the events below, provide their correct order in history. Each event is accompanied by a description and the title of the article it comes from. Return the correct chronological order by listing the event numbers, like (2) (1) (3). (1) Event about `Barnstaple Town railway station': Barnstaple Town station was closed. (2) Event about `RTHK TV 31': RTHK TV 31 closed its analogue service. (3) Event about `Monade': Monade disbanded. (4) Event about `Order of the African Star': The Order of the African Star became a Belgian Order. & (4) (1) (3) (2) \\ \hline \hline
RecurrenceQA & The following event includes the article title and event description. Event about `2008 Iranian legislative election': The second round of the 2008 Iranian legislative election was held. Identify the year when the previous edition of the event took place. Provide the year as your answer. & 2004 \\ \hline 
RecurrenceQA & The following event includes the article title and event description. Event about `2024 Green National Convention': The Green Party National Political Convention began. When was the previous edition of this event? Provide the year as your answer. & 2020 \\
\hline
\end{tabular}
\caption{Examples of Question-Answers pairs for SequenceQA and RecurrenceQA question types in English}
\label{tab:questions_english_part3}
\end{table*}

\end{document}